\pdfoutput=1
\documentclass[11pt]{article}
\usepackage{booktabs}
\usepackage[final]{acl}

\usepackage{times}
\usepackage{latexsym}

\usepackage{booktabs,tabularx,enumitem,etoolbox}
\newlist{procon}{itemize}{1}
\setlist[procon]{nosep,leftmargin=*,
  label=--,% long dash bullet
  after=\vspace{-0.35\baselineskip}}    % tighten slightly
\usepackage{tabularx,booktabs,enumitem,xspace}

\newlist{tlist}{itemize}{1}
\setlist[tlist]{leftmargin=*,nosep,label=--,itemsep=0pt,parsep=0pt}

\usepackage{rotating,threeparttable,booktabs}
\usepackage{booktabs,threeparttable,tabularx,adjustbox}
\usepackage{microtype}         % already loaded by acl.sty, kept for completeness
\usepackage{amsmath}

\usepackage{booktabs}   % \toprule …
\usepackage{pifont}     % \ding{51} / \ding{55}
\newcommand{\cmark}{\ding{51}}  % ✔
\newcommand{\xmark}{\ding{55}}  % ✗

\usepackage{multirow}
\usepackage{tcolorbox}
\usepackage{pifont}           % 提供 \ding

\DeclareUnicodeCharacter{2713}{\cmark}
\DeclareUnicodeCharacter{2717}{\xmark}

\usepackage{rotating,threeparttable}

\usepackage{longtable,booktabs,array,siunitx,xcolor}
\usepackage{url}
\usepackage{etoolbox}      % for \newtoggle
\newcolumntype{L}[1]{>{\raggedright\arraybackslash}p{#1}}

\definecolor{catFSR}{HTML}{F0F9FF}   % fingerspelling
\definecolor{catISLR}{HTML}{F6FFF0}  % isolated
\definecolor{catCSLR}{HTML}{FFF9F0}  % continuous
\definecolor{catSLT}{HTML}{FFF0F6}   % translation
\definecolor{catSLP}{HTML}{F2F0FF}   % production
\newcommand{\rowcolorcat}[1]{%
  \ifstrequal{#1}{FSR}{\rowcolor{catFSR}}{}%
  \ifstrequal{#1}{ISLR}{\rowcolor{catISLR}}{}%
  \ifstrequal{#1}{CSLR}{\rowcolor{catCSLR}}{}%
  \ifstrequal{#1}{SLT}{\rowcolor{catSLT}}{}%
  \ifstrequal{#1}{SLP}{\rowcolor{catSLP}}{}%
}

\newcommand{\mkurl}[2]{%
  \urldef\tempurl\url{#2}\addtocounter{footnote}{1}%
  \footnotetext[\value{footnote}]{\tempurl}%
  \textsuperscript{\thefootnote}%
}

\usepackage[T1]{fontenc}
\usepackage[utf8]{inputenc}

\usepackage{microtype}

\usepackage{inconsolata}

\usepackage{graphicx}
\usepackage{enumitem} % For customizing lists (enumerate, itemize)
\usepackage{hyperref} % For clickable URLs
\usepackage[dvipsnames]{xcolor} 
\usepackage[normalem]{ulem}

\usepackage{tabularx} % 确保导言区包含

\usepackage{graphicx}      % 加载图片
\usepackage{caption}       % 控制 caption 样式
\usepackage{subcaption}    % 支持 subfigure 环境（注意不是 subfigure 包）
\usepackage{float}         % 支持 [H] 强制定位
\usepackage{booktabs}      % 如果配合表格使用（如图+表混合）
\usepackage{xcolor}        % 如果你在图注中使用颜色标注

\usepackage{hyperref}
\usepackage{amsmath,amssymb}
\usepackage{rotating}   % 提供 sidewaystable 环境
\usepackage{amssymb}     % \checkmark
\usepackage{textcomp}
\title{Sign-Language Datasets at Scale:~A Comprehensive Survey on Resources, Benchmarks, and Annotation Standards}

\author{
Yiming Ni \quad
Zhi-Qi Cheng\thanks{Corresponding author.} \quad
Jiayu Li \quad
Wei Cheng \\
Tacoma School of Engineering \& Technology, University of Washington \\
\texttt{\{yimingn, zhiqics, jiayu7, uwcheng\}@uw.edu}
}

\begin{document}
\maketitle

\begin{abstract}
Sign languages are expressive visual languages used by Deaf and Hard-of-Hearing (DHH) communities. 
Despite substantial progress in sign-language recognition, translation, and production, advances remain constrained by fragmented datasets, inconsistent annotations, and limited linguistic coverage. 
Existing benchmarks often fail to reflect real-world communication needs, and systematic analyses of these limitations remain limited.
In this survey, we present a comprehensive index of sign-language datasets, covering \num{120} resources across \num{35} sign languages. 
We analyze key challenges such as modality imbalance, annotation granularity, and signer bias, and outline considerations for future dataset design. 
We also introduce a 24-field \emph{Sign-Language Datasheet} and release a public GitHub repository~\footnote{\url{https://github.com/Ginqwerty/Open-Sign-Language}} to support standardized documentation and reproducible evaluation.
Overall, our work provides a unified and practical foundation for developing inclusive, robust, and scalable sign-language technologies in real-world applications.
\end{abstract}

\section{Introduction}
\label{sec:intro}
Sign languages are fully developed visual–gestural languages used by Deaf and Hard-of-Hearing (DHH) communities, with over 70 million users worldwide~\cite{who_deafness_WHO}. 
Unlike spoken languages, they convey meaning through coordinated manual articulations such as handshape, location, movement, and orientation, together with non-manual signals including facial expressions, mouthing, gaze, and body posture~\cite{boyes-braem_sutton-spence_2001}. 
Despite their linguistic richness~\cite{jachova2008differences}, sign languages remain difficult for hearing populations to acquire, and fluency outside DHH communities is still limited~\cite{kemp1998asl_challenge}. 
This gap continues to create challenges for effective communication between DHH and hearing individuals.

\begin{figure}[t]
\centering
\includegraphics[width=\linewidth]{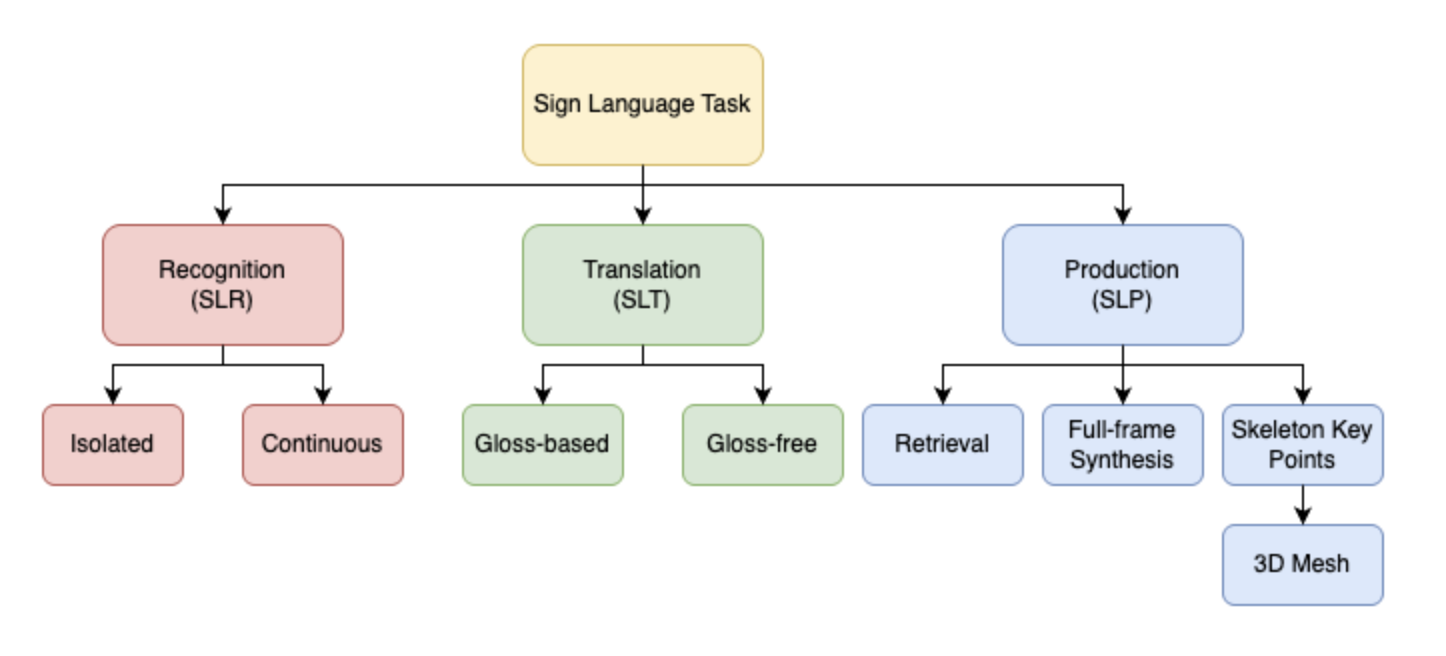}
\vspace{-0.3in}
\caption{\small Overview of sign language tasks: Recognition (SLR), Translation (SLT), and Production (SLP), with representative subtypes and annotation settings.}
\label{fig:taxonomy}
\vspace{-0.2in}
\end{figure}

Human interpreters help bridge this gap, but access is often limited by availability, cost, and scheduling constraints~\cite{universal_translation_services_interpreter_cost}. 
These limitations have motivated increased interest in automated sign language technologies. 
Recent work covers three main tasks, namely recognition, translation, and production, as illustrated in Figure~\ref{fig:taxonomy}. 
However, progress remains closely tied to the quality and coverage of available datasets. 
In practice, existing datasets are fragmented, annotation schemes vary widely, and cross-lingual coverage remains uneven.

Current research relies heavily on a small number of benchmark datasets, while many available resources still receive little attention. 
Existing surveys tend to focus on individual tasks or limited subsets of datasets, and only a few examine datasets in a systematic way with respect to diversity, annotation design, and task suitability. 
As a result, the overall dataset landscape is not yet clearly understood, which significantly limits the development of more robust and generalizable methods. 
More fundamentally, current model designs are often constrained by dataset properties rather than task requirements, highlighting the need for a more data-centric perspective on sign language research.

\begin{table*}[t]
  \centering
  \footnotesize
  \setlength{\tabcolsep}{4.5pt}
  \renewcommand{\arraystretch}{1.05}
  \caption{\small
    Comparison of existing survey papers on sign language technology. 
    “Perf.~Eval.” denotes whether the paper includes performance benchmarking. 
    “Std.~\&~Annot.” indicates discussion of dataset standardization or annotation frameworks.
  }
  \label{tab:comparison}
  \vspace{-0.8em}
  \resizebox{\textwidth}{!}{%
  \begin{tabular}{@{}lccccccc@{}}
    \toprule
    \textbf{Survey Paper} &
    \textbf{Survey Category} &
    \textbf{Datasets Covered} &
    \textbf{Dataset Analysis} &
    \textbf{Challenge Analysis} &
    \textbf{Perf.~Eval.} &
    \textbf{Std.~\& Annot.} &
    \textbf{Task Coverage} \\
    \midrule
    % Alyami et al., 2024~\cite{alyami2024reviewing} & SLR & 17  & \xmark & \xmark & \xmark & \xmark & Only SLR \\
    ~\citealt{alyami2024reviewing} & SLR & 17  & \xmark & \xmark & \xmark & \xmark & Only SLR \\
    % Tao et al., 2024~\cite{tao2024sign} & SLR & 24  & \cmark & \xmark & \xmark & \xmark & Only SLR \\
    ~\citealt{tao2024sign} & SLR & 24  & \cmark & \xmark & \xmark & \xmark & Only SLR \\
    % Sarhan \& Frintrop, 2023~\cite{sarhan2023unraveling} & SLR & 8   & \cmark & \cmark & \xmark & \xmark & Only SLR \\
    ~\citealt{sarhan2023unraveling} & SLR & 8   & \cmark & \cmark & \xmark & \xmark & Only SLR \\
    % Minu et al., 2023~\cite{minu2023extensive} & SLR & 16  & \xmark & \xmark & \xmark & \xmark & Only SLR \\
    ~\citealt{minu2023extensive} & SLR & 16  & \xmark & \xmark & \xmark & \xmark & Only SLR \\
    % Madhiarasan \& Roy, 2022~\cite{madhiarasan2022comprehensive} & SLR & 34  & \cmark & \cmark & \xmark & \xmark & Only SLR \\
    ~\citealt{madhiarasan2022comprehensive} & SLR & 34  & \cmark & \cmark & \xmark & \xmark & Only SLR \\
    % Liang et al., 2023~\cite{liang2023sign} & SLT & 15  & \xmark & \cmark & \cmark & \xmark & Only SLT \\
    ~\citealt{liang2023sign} & SLT & 15  & \xmark & \cmark & \cmark & \xmark & Only SLT \\
    % Núñez-Marcos et al., 2023~\cite{nunez2023survey} & SLT & 33  & \cmark & \cmark & \cmark & \xmark & Only SLT \\
    ~\citealt{nunez2023survey} & SLT & 33  & \cmark & \cmark & \cmark & \xmark & Only SLT \\
    % Kumar Attar et al., 2023~\cite{kumar2023state} & SLT & 22  & \cmark & \cmark & \cmark & \xmark & Only SLT \\
    ~\citealt{kumar2023state} & SLT & 22  & \cmark & \cmark & \cmark & \xmark & Only SLT \\
    % Kahlon \& Singh, 2023~\cite{kahlon2023machine} & SLT & 13  & \xmark & \cmark & \xmark & \xmark & Only SLT \\
    ~\citealt{kahlon2023machine} & SLT & 13  & \xmark & \cmark & \xmark & \xmark & Only SLT \\
    % Rastgoo et al., 2024~\cite{rastgoo2024survey} & SLP & 9   & \cmark & \cmark & \cmark & \xmark & Only SLP \\
    ~\citealt{rastgoo2024survey} & SLP & 9   & \cmark & \cmark & \cmark & \xmark & Only SLP \\
    % Tan et al., 2024~\cite{tan2024review} & SLR, SLT, SLP & 25  & \cmark & \cmark & \cmark & \xmark & Partial \\
    ~\citealt{tan2024review} & SLR, SLT, SLP & 25  & \cmark & \cmark & \cmark & \xmark & Partial \\
    % Papastratis et al., 2021~\cite{papastratis2021artificial} & SLR, SLT, SLP & 13  & \cmark & \cmark & \cmark & \xmark & Partial \\
    ~\citealt{papastratis2021artificial} & SLR, SLT, SLP & 13  & \cmark & \cmark & \cmark & \xmark & Partial \\
    % De Sisto et al., 2022~\cite{de-sisto-etal-2022-challenges} & SLR, SLT & 13  & \cmark & \cmark & \cmark & \cmark & No Task Focus \\
    ~\citealt{de-sisto-etal-2022-challenges} & SLR, SLT & 13  & \cmark & \cmark & \cmark & \cmark & No Task Focus \\
    \textbf{Ours} & \textbf{SLR, SLT, SLP} & \textbf{120} & \cmark & \cmark & \cmark & \cmark & \textbf{Complete} \\
    \bottomrule
  \end{tabular}
  }
  \vspace{-0.15in}
\end{table*}

\noindent \textbf{Scope of Survey.}
This paper presents a dataset-oriented survey of 120 publicly documented sign language datasets spanning 35 languages and the three core tasks of SLR, SLT, and SLP. 
We examine key properties, including data modality, signer demographics, and vocabulary scale, and use these to highlight recurring issues such as modality imbalance, annotation inconsistency, and limited generalizability. 
We further introduce a 24-field \emph{Sign-Language Datasheet} for structured documentation, and release a public GitHub repository together with consolidated benchmark results to support transparent reporting and reproducible research. 
A well-designed dataset, in this context, should balance coverage, consistency, accessibility, and alignment with downstream tasks.

\noindent \textbf{Contributions.}
(1) We compile and organize 120 datasets across 35 sign languages and three core tasks. 
(2) We analyze dataset-level challenges, including modality imbalance, signer bias, and annotation inconsistency. 
(3) We provide practical guidelines for dataset construction and documentation, including the proposed datasheet framework. 
(4) We present consolidated benchmark results to facilitate comparison across datasets and tasks.
\section{Background}
\label{sec:background}
We review the linguistic foundations, task taxonomy, and historical evolution of sign language processing to contextualize dataset-centric analysis and benchmarking in later sections.

\noindent \textbf{Linguistic Foundations}~~Sign languages are natural visual–gestural languages comprising two channels:
~(i)~\textit{manual} (handshape, location, movement, orientation) and
~(ii)~\textit{non-manual} (facial expressions, mouthing, gaze, posture)
~\cite{boyes-braem_sutton-spence_2001}.
These asynchronous, multimodal signals challenge conventional sequential modeling paradigms.
As most sign languages lack standardized orthographies, datasets rely on proxy intermediate representations,
most commonly \textit{glosses}, which map signs to approximate spoken-language words.
A smaller subset of datasets adopts phonological encodings (e.g.,~\textsc{HamNoSys}), capturing fine-grained articulatory structure at substantial annotation cost. 
Together, these linguistic and representational constraints shape task formulation and evaluation.

\noindent \textbf{Task Taxonomy}~~Sign language processing spans three core tasks, each with variants that shape dataset design, annotation schemes,
and modeling strategies (see Figure~\ref{fig:taxonomy}):
~(1)~\textbf{Sign Language Recognition (SLR)} predicts gloss sequences from video.
It includes \textit{isolated} SLR~\cite{laines2023isolated,vazquez2021isolated},
where each video contains a single sign, and \textit{continuous} SLR~\cite{gan2024signgraph,zhou2021signbert},
which transcribes unsegmented sign streams.
~(2)~\textbf{Sign Language Translation (SLT)} maps sign videos to spoken-language text.
Early work relied on gloss-based pipelines~\cite{camgoz2020sign,fu2023token,yin2020better};
more recent approaches adopt gloss-free formulations~\cite{Gong_2024_CVPR,guan2024multi,hu2023signbert,chen2022two} that enable direct video-to-text mapping.
~(3)~\textbf{Sign Language Production (SLP)} synthesizes sign videos from text or gloss input, via retrieval-based methods~\cite{saunders2020progressive}, keypoint-based generation~\cite{qi2024signgen},
or full-frame video synthesis~\cite{zuo2024simple, yin2024t2sgptdynamicvectorquantization}.

\begin{table*}[t]
  \centering
  \footnotesize                        % 相较 \scriptsize 更清晰
  \setlength{\tabcolsep}{4.5pt}        % 轻微压缩列间距
  \renewcommand{\arraystretch}{1.05}   % 行距微调
  \caption{\small Concise overview of representative \textit{fingerspelling} datasets.  
  Abbreviations: ASL---American\,SL; ArSL---Arabic\,SL; AzSL---Azerbaijani\,SL; ISL--- Irish\,SL.  
  For the complete list, please refer to our GitHub.}
  \label{tab:fingerspelling_datasets}
  \vspace{-0.1in}
  \resizebox{\textwidth}{!}{%
  \begin{tabular}{@{}lcccccc@{}}
    \toprule
    \textbf{Dataset} & \textbf{Year} & \textbf{Language} & \textbf{\#Signs} & \textbf{\#Samples} & \textbf{\#Signers} & \textbf{Domain}\\
    \midrule
    \textit{ChicagoFSWild}~\cite{shi2018american}          
        & 2018 & ASL & 31 & 7{,}304\,seq.               & 168 & Letters, Chars.\\
    \textit{ASL Digits}~\cite{mavi2020new}                 
        & 2020 & ASL & 10 & 21{,}800\,img.              & 218 & Letters\\
    \textit{ArASL}~\cite{latif2019arasl}                   
        & 2019 & ArSL& 32 & 54{,}049\,img.              & 40  & Letters\\
    \textit{AzSLD Fingerspelling}~\cite{alishzade2025azsld}
        & 2023 & AzSL& 32 & 10{,}864\,img., 3{,}587\,vid.& 43  & Letters\\
    \textit{ISL-HS}~\cite{oliveira2017dataset}             
        & 2017 & ISL & 23 & 468\,vid., 58{,}114\,img.    & 6   & Letters\\
    \bottomrule
  \end{tabular}}
  %\vspace{-0.1in}
\end{table*}

\begin{table*}[t]
  \centering
  \footnotesize                        % 相较 \scriptsize 更清晰
  \setlength{\tabcolsep}{4.5pt}        % 轻微压缩列间距
  \renewcommand{\arraystretch}{1.05}   % 行距微调
  \caption{\small Representative \textit{isolated} sign-language datasets.  
  Abbreviations: ASL—American\,SL; LSFB—Belgian French SL; CSL—Chinese SL; Auslan—Australian SL;  
  LSA—Argentinian SL; TSL—Turkish SL. The full list is available on GitHub.}
  \label{tab:ISL_table}
  \vspace{-0.1in}
  \resizebox{\textwidth}{!}{%
  \begin{tabular}{@{}lccccccc@{}}
    \toprule
    \textbf{Dataset} & \textbf{Year} & \textbf{Lang.} & \textbf{\#Signs} & \textbf{Dur.} & \textbf{\#Samples} & \textbf{\#Signers} & \textbf{Domain}\\
    \midrule
    \textit{MS-ASL}~\cite{joze2018ms}             
        & 2018 & ASL   & 1,000 & $\sim$25 h     & 25,513 vid. & 222 & General\\
    \textit{WLASL}~\cite{li2020WLASL}             
        & 2019 & ASL   & 2,000 & $\sim$14 h     & 21,083 vid. & 119 & General\\
    \textit{ASL Citizen}~\cite{desai2024asl}      
        & 2023 & ASL   & 2,731 & —              & 83,399 vid. & 52  & General\\
    \textit{LSFB-isol}~\cite{fink2021lsfb}        
        & 2021 & LSFB  & 395   & —              & 47,551 vid. & 85  & General\\
    \textit{DEVISIGN}~\cite{chai2014devisign}     
        & 2014 & CSL   & 4,414 & —              & 331,050 vid.& 30  & General\\
    \textit{SLR500}~\cite{huang2018attention}     
        & 2018 & CSL   & 500   & —              & 125,000 vid.& 50  & General\\
    \textit{NMFs-CSL}~\cite{hu2021global}         
        & 2020 & CSL   & 1,067 & —              & 32,010 vid. & 10  & General\\
    \textit{MM-WLAuslan}~\cite{shen2024mm}        
        & 2024 & Auslan& 3,215 & $\sim$2,500 h  & 282,900 vid.& 73  & General\\
    \textit{LSA-64}~\cite{ronchetti2023lsa64}     
        & 2016 & LSA   & 64    & —              & 3,200 vid.  & 10  & Dictionary\\
    \textit{BosphorusSign22k}~\cite{ozdemir2020bosphorussign22k} 
        & 2020 & TSL   & 744   & $\sim$19 h     & 22,542 vid. & 6   & Health/Finance\\
    \textit{AUTSL}~\cite{sincan2020autsl}         
        & 2020 & TSL   & 226   & 21 h           & 38,336 samples & 43 & General\\
    \bottomrule
  \end{tabular}}
  \vspace{-0.1in}
\end{table*}

\noindent \textbf{Task Evolution \& Research Trends}~~
Research has progressed from finger-spelling and isolated sign recognition~\cite{dreuw2007speech,zhou2021signbert}
to sentence-level translation and full video synthesis.
However, progress remains concentrated on a small set of high-resource languages, notably ASL, BSL, CSL, and DGS, leaving many sign languages underrepresented.
SLR has evolved toward continuous settings, introducing challenges such as coarticulation and temporal ambiguity~\cite{hu2023signbert,gan2024signgraph}.
SLT has shifted from gloss-based pipelines to end-to-end architectures, despite persistent data scarcity.
SLP has transitioned from retrieval-based systems to generative models with signer-aware outputs~\cite{saunders2022signing}.
Despite these advances, prior surveys often focus on individual tasks and provide limited analysis of dataset coverage, annotation granularity, or evaluation standards (Table~\ref{tab:comparison}).
By contrast, we present a unified review of \num{120} datasets across SLR, SLT, and SLP, offering systematic insights into modality, annotation depth, linguistic diversity, and task alignment.
Collectively, these trends highlight the need for inclusive and well-documented datasets, which we address through a comprehensive analysis of datasets (Section~\ref{sec:datasets}), benchmark aggregation (Section~\ref{sec:benchmarks}),
and best-practice guidelines for dataset development (Section~\ref{sec:challenge},~\ref{sec:future-dataset}).

\begin{table*}[t]
  \centering
  \footnotesize                          % 比 \scriptsize 更清晰
  \setlength{\tabcolsep}{4.5pt}          % 轻微压缩列间距
  \renewcommand{\arraystretch}{1.05}     % 行距微调
  \caption{\small Representative \textit{continuous} sign-language corpora.  
  Abbreviations: ASL—American SL; BSL—British SL; CSL—Chinese SL;  
  DGS—German SL; Auslan—Australian SL; LSA—Argentinian SL.  
  The full list is available in GitHub repo.}
  \label{tab:continuous_datasets}
  \vspace{-0.1in}
  \resizebox{\textwidth}{!}{%
  \begin{tabular}{@{}lccccccc@{}}
    \toprule
    \textbf{Corpus} & \textbf{Year} & \textbf{Lang.} & \textbf{\#Vocab} & \textbf{Dur.} & \textbf{\#Samples} & \textbf{\#Signers} & \textbf{Domain}\\
    \midrule
    \textit{RWTH-Boston-104}~\cite{dreuw2007speech}          
        & 2007 & ASL    & 104     & 8.7 min        & 201 sents.       & 3             & General\\
    \textit{How2Sign}~\cite{duarte2021how2sign}              
        & 2020 & ASL    & 16k     & 79 h           & 36,783 sents.    & 11            & General\\
    \textit{OpenASL}~\cite{shi2205open}                      
        & 2022 & ASL    & 33k     & 288 h          & —                & $\sim$220     & General\\
    \textit{YouTube-ASL}~\cite{uthus2024youtube}             
        & 2023 & ASL    & 60k     & $\sim$1,000 h  & —                & $>$2,500      & General\\
    \textit{DailyMoth-70 h}~\cite{rust2024towards}           
        & 2024 & ASL    & 19,694  & 75.8 h         & 48,386 clips     & 1             & News\\
    \textit{BSL-1K}~\cite{albanie2020bsl}                    
        & 2020 & BSL    & 1,064   & $\sim$1,000 h  & 273,000 sents.   & 40            & General\\
    \textit{BOBSL}~\cite{albanie2021BBC_Oxford}              
        & 2021 & BSL    & 2,281   & 1,467 h        & 1.2M seq.        & 39            & General\\
    \textit{CSL-Daily}~\cite{zhou2021CSLDaily}               
        & 2021 & CSL    & 2,000   & —              & 20,645 vid.      & 10            & General\\
    \textit{RWTH-PHOENIX14T}~\cite{camgoz2018neural}         
        & 2020 & DGS    & 2,887   & $\sim$10.5 h   & 8,257 sents.     & 9             & Weather\\
    \textit{Auslan-Daily Comm.}~\cite{shen2024auslan}        
        & 2024 & Auslan & 3,064   & —              & 14,041 sents.    & 49            & Daily\\
    \textit{PHOENIX-News}~\cite{yin2024t2sgptdynamicvectorquantization} 
        & 2024 & DGS    & 190k    & 486 h          & —                & 11            & News\\
    \textit{LSA-T}~\cite{dal2022lsa}                         
        & 2022 & LSA-ES & 14,239  & 21.8 h         & 14,880 sents.    & 103           & General\\
    \bottomrule
  \end{tabular}}
  %\vspace{-0.1in}
\end{table*}

\begin{table*}[t]
  \centering
  \footnotesize
  \setlength{\tabcolsep}{5pt}
  \renewcommand{\arraystretch}{1.05}
\caption{\small
    Annotation layers included in today’s most-used continuous sign language corpora.  
    A \cmark\ indicates the layer is provided; a \xmark\ means it is absent.  
    “Multimodal” refers to any additional stream beyond RGB video (e.g., depth, pose skeleton, 3D mesh).  
    A complete inventory of corpora and their metadata is available in our GitHub repository.}
  \label{tab:dataset_annotation_layers}
  \vspace{-0.1in}
  \resizebox{\textwidth}{!}{%
  \begin{tabular}{@{}lccccccc@{}}
    \toprule
    \textbf{Corpus} & \textbf{Lang.} & \textbf{Video} & \textbf{Clip ID} & \textbf{Gloss} & \textbf{Sent. Align.} & \textbf{Multimodal} & \textbf{File Format}\\
    \midrule
    \textit{PHOENIX14T}~\cite{camgoz2018neural} & DGS   & \cmark & \cmark & \cmark & \cmark & \cmark & CSV\\
    \textit{CSL‐Daily}~\cite{zhou2021CSLDaily}   & CSL   & \cmark & \cmark & \cmark & \cmark & \cmark & TXT\\
    \textit{How2Sign}~\cite{duarte2021how2sign}  & ASL   & \cmark & \cmark & \xmark & \cmark & \cmark & CSV\\
    \textit{YouTube‐ASL}~\cite{uthus2024youtube} & ASL   & \xmark & \cmark & \xmark & \cmark & \xmark & TXT\\
    \textit{OpenASL}~\cite{shi2205open}          & Multi & \xmark & \cmark & \cmark & \cmark & \xmark & TSV\\
    \bottomrule
  \end{tabular}}
  \vspace{-0.1in}
\end{table*}

\begin{table*}[t]
  \centering
  \scriptsize
  \setlength{\tabcolsep}{3.5pt}
  \renewcommand{\arraystretch}{0.98}
  \caption{\small \textbf{Positioning the flagship continuous-sign corpora.}
           “Tasks” = which benchmark(s) the field mainly uses the corpus for.
           Abbreviations: SLR–recognition, SLT–translation, SLP–production.}
  \label{tab:corpus-snapshot}
  \vspace{-0.1in}
  \begin{tabularx}{\linewidth}{@{}l >{\raggedright\arraybackslash}p{4.85cm}
                                >{\raggedright\arraybackslash}p{4.85cm} c@{}}
    \toprule
    \textbf{Corpus} & \textbf{Why you \emph{do} want it} &
    \textbf{Why you \emph{don’t}} & \textbf{Tasks} \\
    \midrule
    \textit{PHOENIX14T}~\cite{camgoz2018neural} &
    \begin{tabular}[t]{@{}l@{}}-- CC-BY; effortless download \\
      -- Text-aligned glosses $\rightarrow$ easy SLT baselines \end{tabular} &
    \begin{tabular}[t]{@{}l@{}}-- Only $\approx$10 h train $\Rightarrow$ over-fit risk \\
      -- Weather broadcast domain $\Rightarrow$ narrow vocab \end{tabular} &
    SLR, SLT, SLP \\

    \textit{CSL-Daily}~\cite{zhou2021CSLDaily} &
    \begin{tabular}[t]{@{}l@{}}-- 2k everyday signs (+ depth, skeleton) \\
      -- Signer-independent split shipped \end{tabular} &
    \begin{tabular}[t]{@{}l@{}}-- NDA gate; lab footage $\Rightarrow$ low background variety \\
      -- Light gloss noise \end{tabular} &
    SLR, SLT \\

    \textit{How2Sign}~\cite{duarte2021how2sign} &
    \begin{tabular}[t]{@{}l@{}}-- 79 h RGB + depth + 3-D mesh \\
      -- 3-D avatar drives SLP research \end{tabular} &
    \begin{tabular}[t]{@{}l@{}}-- No manual gloss layer \\
      -- 3 TB raw download $\Rightarrow$ storage heavy \end{tabular} &
    SLT, SLP \\

    \textit{YouTube-ASL}~\cite{uthus2024youtube} &
    \begin{tabular}[t]{@{}l@{}}-- $\approx$1,000 h in-the-wild clips \\
      -- Community can extend corpus on the fly \end{tabular} &
    \begin{tabular}[t]{@{}l@{}}-- Only YT IDs (link-rot, geo-blocks) \\
      -- Heterogeneous quality; no pose/depth \end{tabular} &
    SLT (large-scale pre-train) \\

    \textit{OpenASL}~\cite{shi2205open} &
    \begin{tabular}[t]{@{}l@{}}-- Apache-2.0 TSV annotations \\
      -- 33k open-domain vocab---rare for ASL \end{tabular} &
    \begin{tabular}[t]{@{}l@{}}-- Must crawl videos yourself \\
      -- Mixed gloss standards; tooling scant \end{tabular} &
    SLT (open-domain) \\
    \bottomrule
  \end{tabularx}
  %\vspace{-0.1in}
\end{table*}

\section{Dataset Compendium}
\label{sec:datasets}
High-quality sign language datasets are fundamental to the development of robust models for recognition, translation, and production tasks. 
We organize existing datasets into three main categories:
(i)~\textit{Fingerspelling} datasets, which consist of static images or short video clips of manual alphabets;
(ii)~\textit{Isolated Sign Language Datasets (ISLD)}, where individual signs are recorded as separate video samples; and
(iii)~\textit{Continuous Sign Language Datasets (CSLD)}, which contain longer, continuous sign sequences.
Representative datasets are summarized in Tables~\ref{tab:fingerspelling_datasets}, \ref{tab:ISL_table}, and \ref{tab:continuous_datasets}. 
Complete listings and extended metadata are available in the accompanying public GitHub repository for reference.

\noindent \textbf{Fingerspelling Datasets}~Table~\ref{tab:fingerspelling_datasets} summarizes representative fingerspelling datasets across a range of sign languages, from early, small-scale laboratory benchmarks (e.g., \textit{ASL Digits}~\cite{mavi2020new}, \textit{ArASL}~\cite{latif2019arasl}) to more recent in-the-wild corpora such as \textit{ChicagoFSWild}~\cite{shi2018american} and \textit{AzSLD Fingerspelling}~\cite{alishzade2025azsld}. 
Early datasets are typically collected in controlled settings, but exhibit limited variation in lighting, background, signer demographics, and handshape complexity. 
More recent datasets emphasize greater diversity in participants, higher spatial resolution, and more varied real-world recording conditions, supporting the development of more robust recognition models. 
In addition, broader coverage of larger manual alphabets, including diacritics (e.g., AzSLD~\cite{alishzade2025azsld}), further facilitates cross-lingual transfer and adaptation across sign languages.

\noindent \textbf{Isolated Sign Language Datasets}~Table~\ref{tab:ISL_table} summarizes representative datasets for single-sign recognition. 
Foundational benchmarks such as \textit{MS-ASL}~\cite{joze2018ms} and \textit{WLASL}~\cite{li2020WLASL} introduced medium-to-large vocabularies (approximately 1k--2k signs) and remain widely used due to their signer diversity and broad task coverage. 
More recent datasets extend both vocabulary scale and linguistic coverage. For example, \textit{DEVISIGN}~\cite{chai2014devisign} includes over 300k Chinese Sign Language samples, while \textit{MM-WLAuslan}~\cite{shen2024mm} provides multi-view Auslan recordings that capture greater variation across signers. 
In addition to raw video, newer datasets increasingly incorporate crowd-sourced data and multimodal signals, such as RGB, depth, and skeletal representations, to better capture fine-grained signing behavior. 
Together, these datasets support recent advances in signer-independent recognition, large-vocabulary classification, and multimodal modeling tasks.

\noindent \textbf{Continuous Sign Language Datasets}~~Compared to isolated datasets, Continuous Sign Language Datasets (CSLDs) feature longer, discourse-level signing sequences.
Early examples such as \textit{RWTH-Boston-104}~\cite{dreuw2007speech} contained limited annotated material, while more recent corpora such as \textit{How2Sign}~\cite{duarte2021how2sign} and
\textit{YouTube-ASL}~\cite{uthus2024youtube} scale to hundreds of hours and tens of thousands of unique signs.
These large-scale datasets enable research on continuous sign language recognition (CSLR), translation (SLT), and sign language production (SLP).
Modern CSLDs increasingly provide rich, multi-level annotations (e.g., glosses and sentence alignments), enabling more detailed linguistic analyses of coarticulation, sign boundaries, and domain-specific expressions.
They support the study of spontaneous signing styles, non-manual cues such as facial expressions, and domain variation (e.g., news and conversation).
Effectively leveraging such corpora requires addressing challenges in temporal alignment,
segmentation, and multimodal integration.

\begin{table*}[htbp]
    \centering
    \footnotesize
    \setlength{\tabcolsep}{5pt}
    \renewcommand{\arraystretch}{1.05}
    \caption{\small \textbf{CSLR leaderboard performance} on PHOENIX14T and CSL-Daily. All numbers are word error rates (WER), where lower values indicate better recognition accuracy. Full dataset statistics and links are available at the GitHub repository.}
    \vspace{-0.1in}
    \resizebox{\textwidth}{!}{%
    \begin{tabular}{l c c | l c c}
        \toprule
        \multicolumn{3}{c|}{\textbf{PHOENIX14T}} & \multicolumn{3}{c}{\textbf{CSL-Daily}} \\
        \cmidrule(r){1-3} \cmidrule(l){4-6}
        \textbf{Model} & \textbf{WER ↓} & \textbf{Input} & \textbf{Model} & \textbf{WER ↓} & \textbf{Input} \\
        \midrule
        SignVTCL~\cite{chen2024signvtcl} & 17.9\% & RGB, Skeleton, Flow & SignVTCL~\cite{chen2024signvtcl} & 24.1\% & RGB, Skeleton, Flow \\
        Cross-Ling~\cite{Wei_2023_ICCV} & 18.5\% & RGB & MAM-FSD~\cite{zhu2025continuous} & 24.5\% & RGB \\
        C$^2$ST~\cite{Zhang_2023_ICCV} & 18.9\% & RGB & TwoStream-SLT~\cite{chen2022two} & 25.3\% & RGB, Skeleton \\
        MultiSignGraph~\cite{gan2024signgraph} & 19.1\% & RGB & C$^2$ST~\cite{Zhang_2023_ICCV} & 25.8\% & RGB \\
        TwoStream-SLT~\cite{chen2022two} & 19.3\% & RGB, Skeleton & MultiSignGraph~\cite{gan2024signgraph} & 26.4\% & RGB \\
        \bottomrule
    \end{tabular}
    }
    %\vspace{-0.1in}
    \label{tab:leaderboard_SLR_comparison}
\end{table*}

\begin{table*}[htbp]
    \centering
    \footnotesize
    \setlength{\tabcolsep}{5pt}
    \renewcommand{\arraystretch}{1.05}
    \caption{\small \textbf{Gloss-based SLT leaderboard} on PHOENIX14T and CSL-Daily. BLEU scores are reported on the test set; higher values indicate better translation performance. Full dataset statistics and links are available at the GitHub repository.}
    \vspace{-0.1in}
    \resizebox{\textwidth}{!}{%
    \begin{tabular}{l c c | l c c}
        \toprule
        \multicolumn{3}{c|}{\textbf{PHOENIX14T}} & \multicolumn{3}{c}{\textbf{CSL-Daily}} \\
        \cmidrule(lr){1-3} \cmidrule(lr){4-6}
        \textbf{Model} & \textbf{BLEU ↑} & \textbf{Input} & \textbf{Model} & \textbf{BLEU ↑} & \textbf{Input} \\
        \midrule
        TextCTC-SLT~\cite{tan2024improvement} & 28.42\% & RGB & TwoStream-SLT~\cite{chen2022two} & 25.79\% & RGB, Skeleton \\
        TwoStream-SLT~\cite{chen2022two} & 26.71\% & RGB, Skeleton & SLTUNET~\cite{zhang2023sltunet} & 23.76\% & RGB \\
        SLTUNET~\cite{zhang2023sltunet} & 26.00\% & RGB & TextCTC-SLT~\cite{tan2024improvement} & 22.47\% & RGB \\
        ConSLT~\cite{fu2023token} & 25.48\% & RGB & MMTLB~\cite{Chen_2022_CVPR} & 21.46\% & RGB \\
        MMTLB~\cite{Chen_2022_CVPR} & 24.60\% & RGB & BN-TIN-Transf + BT~\cite{zhou2021CSLDaily} & 19.67\% & RGB \\
        \bottomrule
    \end{tabular}
    }
    \vspace{-0.1in}
    \label{tab:leaderboard_SLT}
\end{table*}

\begin{table*}[htbp]
    \centering
    \footnotesize
    \setlength{\tabcolsep}{5pt}
    \renewcommand{\arraystretch}{1.05}
    \caption{\small \textbf{Gloss-free SLT leaderboard} on PHOENIX14T, CSL-Daily, and How2Sign. BLEU scores are reported on the test set; higher values indicate better translation performance. Full leaderboard details and links are available at the GitHub repository.}
    \vspace{-0.1in}
    \resizebox{\textwidth}{!}{%
    \begin{tabular}{l c | l c | l c}
        \toprule
        \multicolumn{2}{c|}{\textbf{PHOENIX14T}} & \multicolumn{2}{c|}{\textbf{CSL-Daily}} & \multicolumn{2}{c}{\textbf{How2Sign}} \\
        \cmidrule(lr){1-2} \cmidrule(lr){3-4} \cmidrule(lr){5-6}
        \textbf{Model} & \textbf{BLEU ↑} & \textbf{Model} & \textbf{BLEU ↑} & \textbf{Model} & \textbf{BLEU ↑} \\
        \midrule
        CV-SLT~\cite{zhao2024conditional}         & 29.27\% & Uni-Sign~\cite{li2025unisignunifiedsignlanguage} & 26.36\% & SSVP-SLT~\cite{rust2024towards} & 15.5\% \\
        MSKA-SLT~\cite{guan2024multi}             & 29.03\% & MSKA-SLT~\cite{guan2024multi}         & 25.52\% & Uni-Sign~\cite{li2025unisignunifiedsignlanguage} & 14.9\% \\
        TwoStream-SLT~\cite{chen2022two}          & 28.95\% & TwoStream-SLT~\cite{chen2022two}      & 25.42\% & SignMusketeers~\cite{gueuwou2025signmusketeersefficientmultistreamapproach} & 14.3\% \\
        SLTUNET~\cite{zhang2023sltunet}           & 28.47\% & SLTUNET~\cite{zhang2023sltunet}        & 25.01\% & VAP~\cite{jiao2024visual}                          & 12.87\% \\
        MMTLB~\cite{Chen_2022_CVPR}               & 28.39\% & MMTLB~\cite{Chen_2022_CVPR}            & 23.92\% & SLT-CC~\cite{jang2025lost}                 & 12.70\% \\
        IP-SLT~\cite{Yao_2023_ICCV}               & 27.97\% & C$^2$ST~\cite{Zhang_2023_ICCV}          & 21.61\% & YouTube-ASL~\cite{uthus2024youtube}               & 12.39\% \\
        C$^2$RL~\cite{Zhang_2023_ICCV}            & 26.75\% & XmDA~\cite{ye2023cross}               & 21.58\% & SLT-SEM~\cite{hamidullah2024sign}                 & 11.70\% \\
        VAP~\cite{jiao2024visual}                 & 26.16\% & BN-TIN-Transf + BT~\cite{zhou2021CSLDaily} & 21.34\% & FLa-LLM~\cite{chen2024factorized}                 &  9.66\% \\
        \bottomrule
    \end{tabular}
    }
    \label{tab:leaderboard_SLT_gloss_free}
\end{table*}

\begin{table*}[htbp]
    \centering
    \footnotesize
    \setlength{\tabcolsep}{5pt}
    \renewcommand{\arraystretch}{1.05}
    \caption{\small \textbf{SLP leaderboard} for \textbf{Gloss-to-Pose} and \textbf{Text-to-Pose} models. BLEU scores are reported on the test set; higher values indicate better video generation performance. Full dataset details and links are available at the GitHub repository.}
    \vspace{-0.1in}
    \resizebox{\textwidth}{!}{%
    \begin{tabular}{l c | l c c}
        \toprule
        \multicolumn{2}{c|}{\textbf{Gloss-to-Pose}} & \multicolumn{3}{c}{\textbf{Text-to-Pose}} \\
        \cmidrule(lr){1-2} \cmidrule(lr){3-5}
        \textbf{Model} & \textbf{BLEU ↑} & \textbf{Model} & \textbf{BLEU ↑} & \textbf{Gloss-Free} \\
        \midrule
        FS-NET~\cite{saunders2022signing} & 18.78\% & Spoken2Sign~\cite{zuo2024simple} & 25.46\% & No \\
        Adversarial Training~\cite{saunders2020adversarialtrainingmultichannelsign} & 11.70\% & SignDiff~\cite{fang2023signdiff} & 22.15\% & Yes \\
        Progressive Transf~\cite{saunders2020progressive} & 10.43\% & FS-NET~\cite{saunders2022signing} & 21.10\% & Yes \\
        NSLP-G~\cite{hwang2021non} & 9.39\% & SignGen~\cite{qi2024signgen} & 19.71\% & Yes \\
        LVMCN~\cite{wang2024linguistics} & 9.36\% & T2S-GPT~\cite{yin2024t2sgptdynamicvectorquantization} & 11.87\% & Yes \\
        Data-Driven~\cite{walsh2024data} & 9.17\% & NSLP-G (+ Fine-tuning)~\cite{hwang2021non} & 11.07\% & Yes \\
        \bottomrule
    \end{tabular}
    }
    \vspace{-0.1in}
    \label{tab:leaderboard_SLP}
\end{table*}

% \begin{table}[!htbp]
%     \centering
%     \footnotesize
%     \setlength{\tabcolsep}{5pt}
%     \renewcommand{\arraystretch}{1.05}
%     \caption{\small \textbf{Text-to-Pose leaderboard} on PHOENIX14T in the SLP task. Models are evaluated by BLEU on the test set; higher scores indicate better generation quality. ``ft'' denotes models that are fine-tuned from pretrained checkpoints.}
%     \label{tab:leaderboard_SLP_text_only}
%     \vspace{-0.1in}
%     \begin{tabularx}{\linewidth}{@{}p{4.5cm} c c@{}}
%         \toprule
%         \textbf{Model} & \textbf{BLEU ↑} & \textbf{Gloss-Free} \\
%         \midrule
%         Spoken2Sign~\cite{zuo2024simple} & 25.46\% & No \\
%         SignDiff~\cite{fang2023signdiff} & 22.15\% & Yes \\
%         FS-NET~\cite{saunders2022signing} & 21.10\% & Yes \\
%         SignGen~\cite{qi2024signgen} & 19.71\% & Yes \\
%         T2S-GPT~\cite{yin2024t2sgptdynamicvectorquantization} & 11.87\% & Yes \\
%         NSLP-G (ft)~\cite{hwang2021non} & 11.07\% & Yes \\
%         \bottomrule
%     \end{tabularx}
% \end{table}

% 然后在正文里这样写 figure*
\begin{figure*}[!htbp]
  \centering
  % 把整图缩放到 0.9\textwidth，节省篇幅
  \resizebox{1.0\textwidth}{!}{%
    \begin{minipage}{\textwidth}
      \centering
      \begin{subfigure}[t]{0.28\textwidth}
        \includegraphics[width=\linewidth]{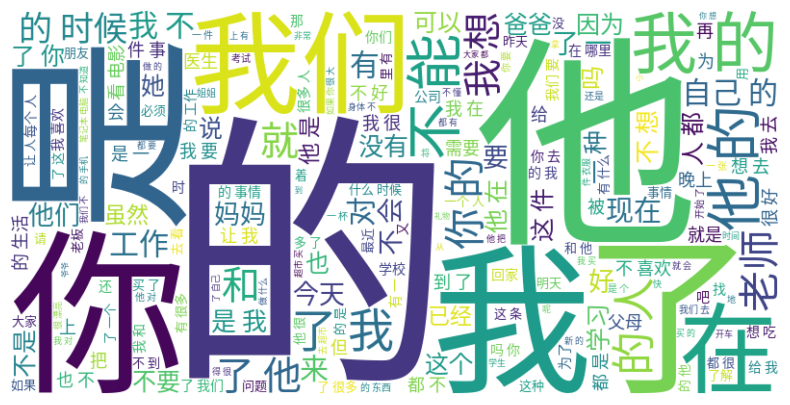}  
        \caption{CSL-Daily}
      \end{subfigure}\hfill
      \begin{subfigure}[t]{0.28\textwidth}
        \includegraphics[width=\linewidth]{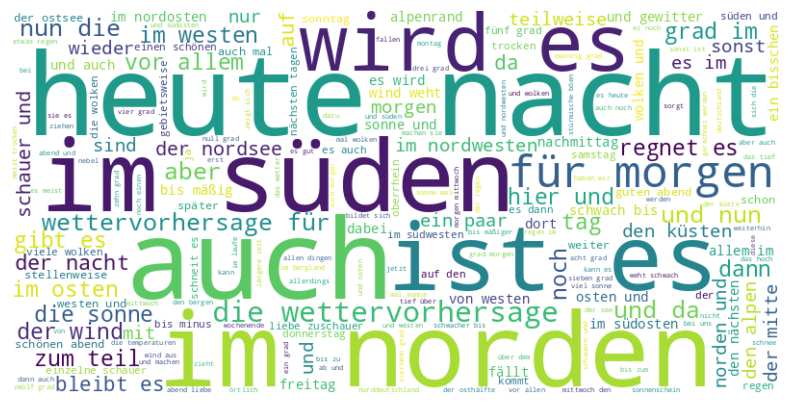}
        \caption{PHOENIX14T}
      \end{subfigure}\hfill
      \begin{subfigure}[t]{0.28\textwidth}
        \includegraphics[width=\linewidth]{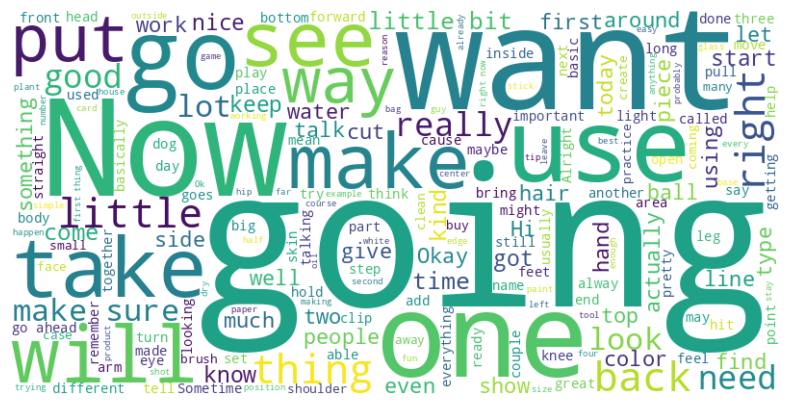}
        \caption{How2Sign}
      \end{subfigure}
    \end{minipage}%
  }% end \resizebox
  \vspace{-0.1in}
  \caption{\small \textbf{Word clouds of translation outputs} from three major SLT datasets: CSL-Daily, PHOENIX14T, and How2Sign. The visualization highlights frequent words in target sentences, revealing domain-specific vocabulary distributions.}
  \label{fig:three_subfigs}
  \vspace{-0.2in}
\end{figure*}

\section{Benchmarks \& Leaderboards}
\label{sec:benchmarks}
Building on the datasets introduced in Section~\ref{sec:datasets}, we conduct a systematic benchmark analysis across sign language recognition, translation, and production.
This section compares the performance of representative models on five widely used benchmark datasets:
PHOENIX14T, CSL-Daily, How2Sign, YouTube-ASL, and OpenASL.
The results are reported by task (SLR, SLT, and SLP) and stratified into gloss-based and gloss-free settings.

\noindent \textbf{Recognition Benchmarks (SLR)}~~Table~\ref{tab:leaderboard_SLR_comparison} reports WER performance of recent models on PHOENIX14T and CSL-Daily. 
PHOENIX14T consistently yields lower error rates, with SignVTCL~\cite{chen2024signvtcl} achieving 17.9\%. 
This can be attributed to its clean annotation pipeline, narrow topical focus (weather domain), and relatively limited signer variation, which together facilitate more stable motion-to-text alignment under controlled conditions. 
In contrast, CSL-Daily exhibits higher WERs (lowest 24.1\%) despite comparable model architectures. 
This reflects its greater diversity in signers, topics, and recording environments, as well as the inclusion of casual daily expressions and multimodal inputs (RGB, depth, and skeleton). 
While these factors increase modeling difficulty, they also improve ecological validity. 
As a result, models show larger performance gaps on CSL-Daily than on PHOENIX14T, highlighting its value as a benchmark for generalization. 
For practical deployment, CSL-Daily provides a more realistic and challenging testbed, particularly for evaluating signer independence, coarticulation effects, and robustness under natural conditions.

\noindent \textbf{Translation Benchmarks (SLT)}~We compare gloss-based and gloss-free SLT on PHOENIX14T, CSL-Daily, and How2Sign, which differ in annotation structure, domain, and linguistic complexity, leading to distinct benchmarking characteristics. 
Under current \textsc{BLEU}-centric evaluation settings and high-resource corpora, systems with intermediate gloss supervision often achieve higher scores. 
However, this advantage is largely driven by supervision availability, domain consistency, and metric sensitivity, rather than an inherent benefit of gloss-based formulations. 
In contrast, gloss-free approaches reduce annotation cost and offer greater scalability, particularly for languages without standardized gloss conventions. 
This trade-off highlights a broader tension between evaluation performance and practical applicability in SLT systems.

\noindent \textbf{Gloss-based SLT}
Table~\ref{tab:leaderboard_SLT} reports BLEU scores for models trained with intermediate gloss supervision. 
PHOENIX14T consistently achieves higher performance, with TextCTC-SLT~\cite{tan2024improvement} reaching 28.42\% BLEU. 
This can be attributed to its relatively narrow domain and well-aligned gloss--sentence pairs, which support more stable learning of structured mappings. 
In contrast, CSL-Daily covers more diverse everyday topics and exhibits greater variation across signers. 
Consequently, BLEU scores are generally lower (up to 25.8\%), but the dataset provides a more realistic and challenging setting for evaluating semantic generalization. 
This comparison highlights an important trade-off between benchmark performance and real-world complexity in gloss-based SLT.

\noindent \textbf{Gloss-free SLT}
Table~\ref{tab:leaderboard_SLT_gloss_free} reports results for end-to-end models that translate sign language videos directly into spoken language without gloss supervision. 
Although gloss-free methods generally achieve lower BLEU scores than gloss-based approaches, they offer improved scalability and substantially reduced annotation cost. 
PHOENIX14T and CSL-Daily remain the primary benchmarks for this setting. 
In contrast, How2Sign yields lower BLEU scores (best: 15.5\%), but its large vocabulary, multi-camera recordings, and absence of gloss annotations make it particularly valuable for evaluating large-scale and real-world scenarios. 
Overall, the performance gap between gloss-based and gloss-free methods has narrowed, reflecting steady progress in end-to-end modeling. 
This trend has shifted recent research toward multimodal pretraining and scaling strategies, especially in low-resource and open-domain settings.

\noindent \textbf{Production Benchmarks (SLP)}~~
We evaluate sign language production (SLP) models that generate sign videos from either gloss inputs (Gloss-to-Pose) or spoken-language text (Text-to-Pose). Table~\ref{tab:leaderboard_SLP} reports \textsc{BLEU} scores for both settings. 
Current SLP research lacks standardized pipelines for pose extraction, 3D lifting, and evaluation, and many models are not publicly available, limiting reproducibility. 
As a result, comparisons across studies are inconsistent, and our analysis is based on reported leaderboard results. 
Among Gloss-to-Pose models, FS-NET~\cite{saunders2022signing} achieves the highest score (18.78\%), benefiting from alignment-aware supervision. 
For Text-to-Pose, Spoken2Sign~\cite{zuo2024simple} attains the best performance (25.46\%), despite the more complex input space, suggesting the effectiveness of large-scale text encoders. 
Other approaches, including SignDiff~\cite{fang2023signdiff} and SignGen~\cite{qi2024signgen}, adopt diffusion-based generative modeling to improve visual realism. 
Overall, performance differences in SLP are influenced by evaluation protocols and dataset design, rather than by a clear advantage of gloss conditioning. 
At the same time, gloss-free text-to-pose methods reduce annotation cost and scale more naturally, with the performance gap continuing to narrow under multimodal conditioning and large-scale pretraining.

\noindent \textbf{Text-only SLP}~~Gloss-free approaches such as SignDiff and SignGen achieve competitive \textsc{BLEU} scores without relying on intermediate gloss annotations. 
Spoken2Sign remains the strongest-performing model, indicating that effective textual pretraining can partially compensate for the absence of explicit gloss structure. 
Additional models, including T2S-GPT~\cite{yin2024t2sgptdynamicvectorquantization} and NSLP-G (+fine-tuning)~\cite{hwang2021non}, further demonstrate the benefits of fine-tuning, although they continue to lag slightly behind the top-performing systems. 
Overall, the field is increasingly moving toward direct Text-to-Pose modeling, which offers improved scalability and reduced annotation requirements. 
However, maintaining visual fidelity and temporal coherence remains a key challenge, particularly in unconstrained real-world settings.

\noindent \textbf{Future Evaluation for SLP}~~SLP remains a relatively new task with limited publicly available work, and current evaluations are largely concentrated on \textsc{PHOENIX-2014T} and How2Sign. 
To maintain comparability, \textsc{BLEU} is commonly reported when back-translation is used; however, its reliability is fundamentally constrained by the underlying SLT model. 
In particular, several How2Sign evaluations~\cite{fang2023signdiff, hwang2024gloss} rely on pre-trained back-translators with undisclosed training details, leading to substantial variation in reported results. 
Accordingly, \textsc{BLEU} should be interpreted primarily as a relative, rather than absolute, measure of generation quality. 

To obtain a more complete assessment of intelligibility and deployability, we suggest complementing \textsc{BLEU} with additional metrics such as MPJPE$_{\text{DTW}}$, Hand-MJE, timing~F1, and human evaluation. 
For reproducibility, experimental settings should also explicitly specify key factors including input modality (RGB, pose, or fusion), supervision type (gloss-conditioned or text-to-pose), use of large-scale pretraining, and sampling rate (fps). 
Such reporting improves comparability while remaining compatible with existing benchmarks.

For image- or video-based SLP, which involves full-frame visual synthesis, evaluation further requires measures of visual realism and temporal coherence beyond pose-based metrics. 
Common choices include perceptual and video-quality metrics such as PSNR, SSIM, and LPIPS, along with distributional measures such as FVD to capture motion diversity and temporal dynamics. 
However, these metrics remain largely insensitive to fine-grained articulation of hands and facial expressions, which are central to sign language. 
We therefore emphasize the need to combine them with motion-aware metrics and human evaluation to ensure both perceptual and linguistic fidelity.

While strict unification across datasets is unlikely, we advocate reporting a consistent set of complementary metrics under clearly specified settings, enabling more reliable and more comparable evaluation across studies. 
More broadly, current evaluation protocols often reflect dataset-specific assumptions and convenience rather than communicative effectiveness in real-world settings.

\begin{figure}[!t]
    \centering
    \includegraphics[width=\linewidth]{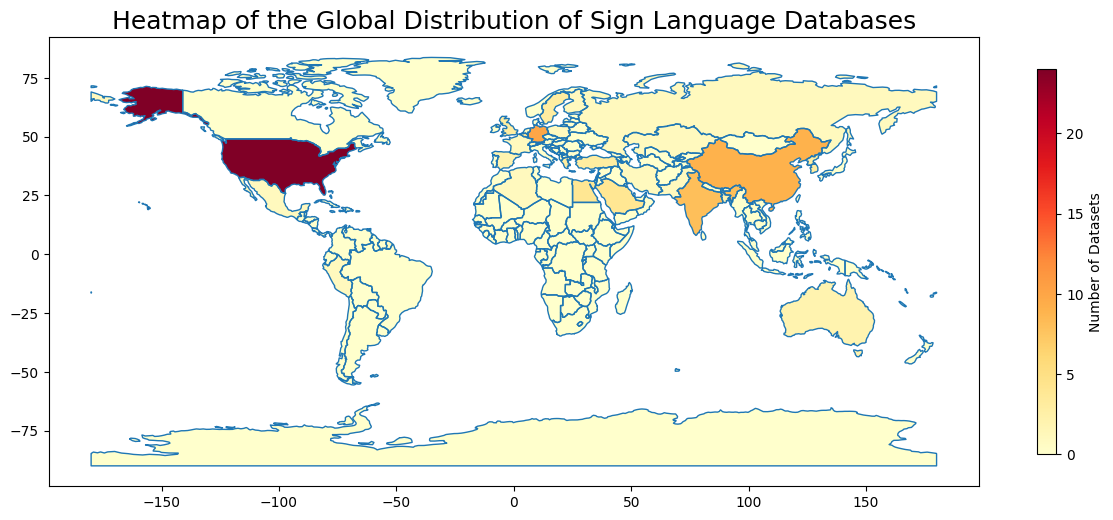}
    \vspace{-0.15in}
    \caption{\small \textbf{Geographic distribution of sign language datasets.} The heatmap highlights the number of datasets collected per country or region. Darker colors indicate higher dataset density, with most resources concentrated in Europe, North America, and East Asia.~[Best zooming in to view].}
    \vspace{-0.2in}
    \label{fig:world_heatmap}
\end{figure}

\begin{figure*}[!ht]
    \centering
    \begin{subfigure}{0.185\textwidth}
        \includegraphics[width=\linewidth]{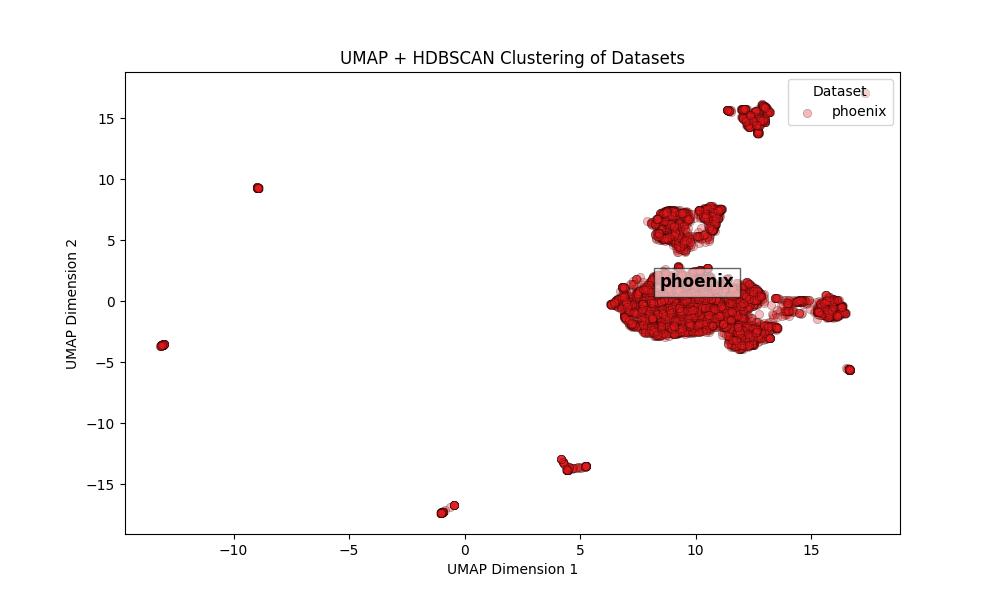}
        \caption{PHOENIX14T}
    \end{subfigure}
    \hfill
    \begin{subfigure}{0.185\textwidth}
        \includegraphics[width=\linewidth]{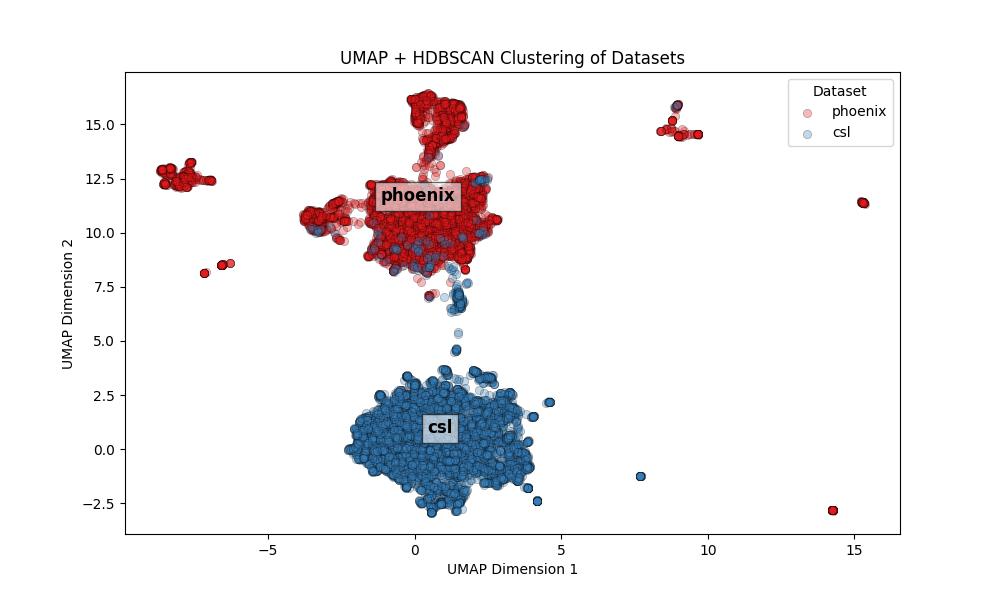}
        \caption{+ CSL-Daily}
    \end{subfigure}
    \hfill
    \begin{subfigure}{0.185\textwidth}
        \includegraphics[width=\linewidth]{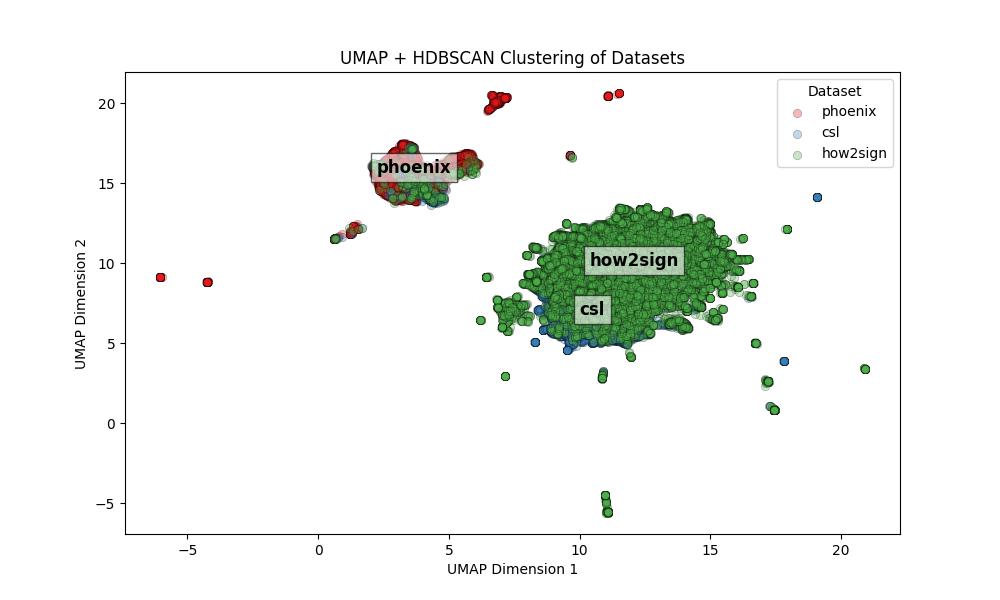}
        \caption{+ How2Sign}
    \end{subfigure}
    \hfill
    \begin{subfigure}{0.185\textwidth}
        \includegraphics[width=\linewidth]{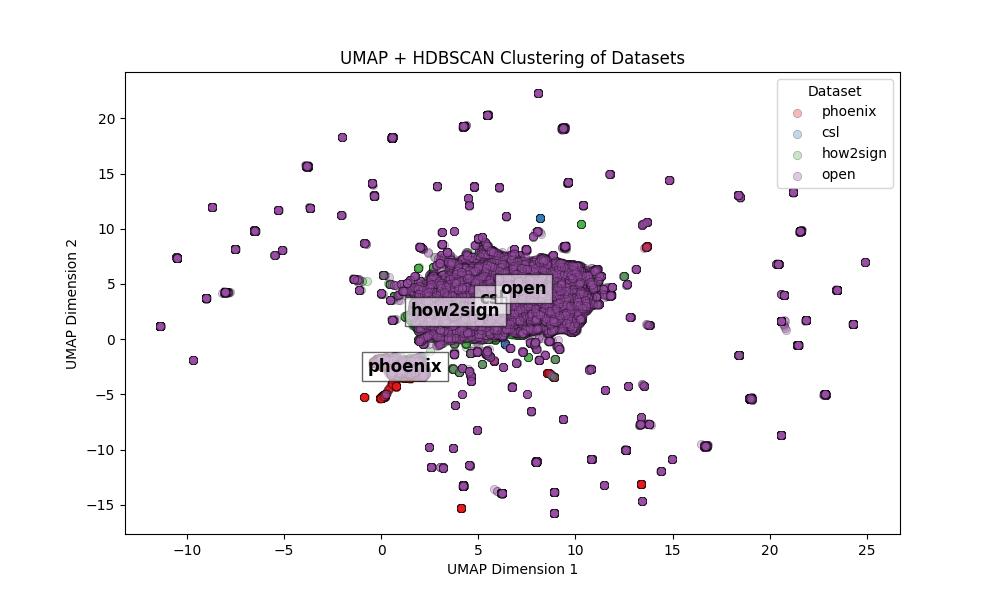}
        \caption{+ OpenASL}
    \end{subfigure}
    \hfill
    \begin{subfigure}{0.185\textwidth}
        \includegraphics[width=\linewidth]{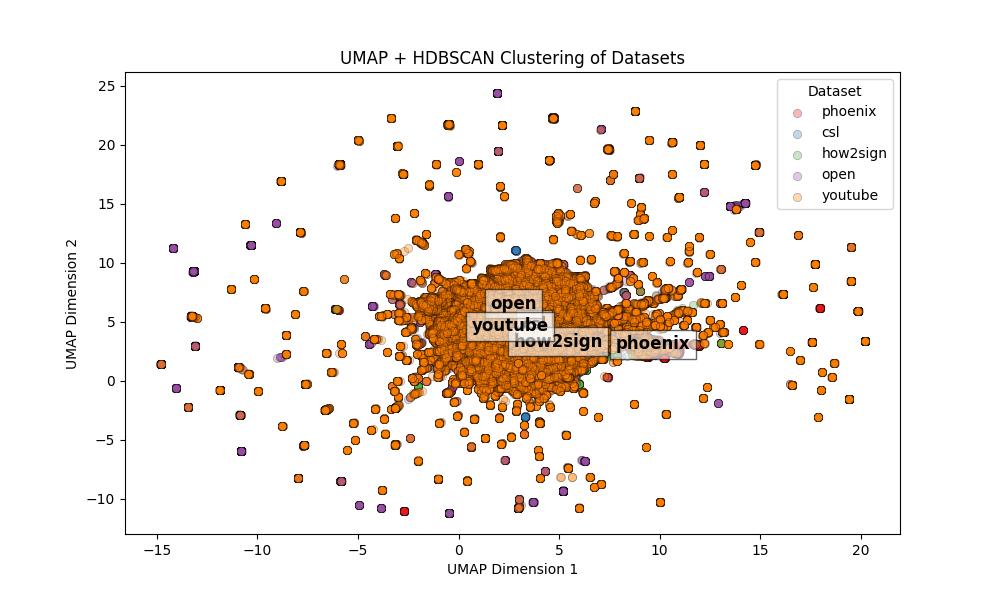}
        \caption{+ YouTube-ASL}
    \end{subfigure}
    \vspace{-0.05in}
    \caption{\small \textbf{UMAP projection of sentence embeddings across datasets.} Each panel incrementally adds one dataset to PHOENIX14T, illustrating how semantic domains expand and overlap in embedding space. Colors: PHOENIX14T (red), CSL-Daily (blue), How2Sign (green), OpenASL (purple), YouTube-ASL (orange).~[Best zooming in to view].}
    \vspace{-0.1in}
    \label{fig:sentence_Embedding_with_UMAP}
\end{figure*}

\section{Dataset Challenges}
\label{sec:challenge}
Despite rapid progress in sign language modeling, several structural challenges remain, particularly in accessibility, linguistic coverage, annotation practices, and ecological validity. 
In this section, we identify five key issues based on the visualizations and benchmark analyses presented earlier. 
Although we attempted to quantify factors such as inter-annotator agreement (IAA), demographic diversity, and ecological validity, our audit shows that these attributes are rarely reported in publicly available corpora. 
We therefore treat such omissions as documentation gaps, rather than attempting to directly infer or impute missing values.

\noindent \textbf{Access Barriers \& Sustainability}
Although more than 100 sign language datasets have been released, only a small subset is widely used. 
Rather than excluding historically important but currently unavailable datasets, we explicitly document their accessibility status. 
As shown in Table~\ref{tab:dataset_annotation_layers} and Table~\ref{tab:corpus-snapshot}, 
datasets such as CSL-Daily and BOBSL require data use agreements or institutional approval, which limits their adoption in open research settings. 
Earlier datasets, including SIGNUM~\cite{von2010signum}, are affected by link rot and are no longer accessible, while resources such as YouTube-ASL provide only video identifiers, making reproducibility fragile and long-term access uncertain. 
In contrast, PHOENIX14T remains widely used due to its open availability, well-aligned gloss annotations, and consistent data format, despite its relatively limited scale. 
Taken together, these observations suggest that accessibility, documentation quality, and long-term availability are key factors shaping dataset impact and sustainability.

\noindent \textbf{Linguistic \& Geographic Imbalance}
Figure~\ref{fig:world_heatmap} shows that publicly available corpora are concentrated in a small set of high-resource sign languages (e.g., ASL, DGS, CSL, and ISL), while many others, particularly those in South Asia, Africa, and Indigenous or village communities, remain largely unrepresented. 
Even within a single language, regional variation is rarely documented, and signer-level attributes are similarly underreported. 
In particular, handedness is seldom recorded, despite its linguistic relevance and the non-trivial prevalence of left-hand dominance in deaf populations. 
The lack of such metadata limits the analysis of signer variability and can introduce biases in model training and evaluation. 
High-resource languages benefit from large, richly annotated corpora (e.g., YouTube-ASL), whereas underrepresented languages often rely on smaller, lab-collected datasets with limited metadata or restricted access, further reinforcing existing disparities. 
Figure~\ref{fig:sentence_Embedding_with_UMAP} shows that sentence embeddings from different datasets (PHOENIX14T, CSL-Daily, How2Sign, OpenASL, and YouTube-ASL) form largely disjoint clusters, indicating weak semantic alignment across domains. 
This fragmentation limits multi-dataset pretraining and reduces the effectiveness of zero-shot transfer.
Taken together, these observations point to a structural mismatch between dataset coverage and real-world linguistic diversity, which remains a central obstacle to building generalizable sign language models.

\noindent \textbf{Inconsistent Modalities \& Annotations}
Sign language datasets vary widely in input modality (RGB, depth, pose), data format (CSV, TSV, JSON), and annotation layers (e.g., glosses and sentence alignment). 
As shown in Table~\ref{tab:dataset_annotation_layers}, only PHOENIX14T and CSL-Daily provide relatively complete supervision, whereas datasets such as OpenASL and YouTube-ASL lack gloss annotations or synchronized modalities. 
This heterogeneity complicates joint modeling and undermines reproducibility. 
Even within individual datasets, annotation conventions are not standardized; for example, translation fields are labeled \texttt{translation} in PHOENIX14T but \texttt{SENTENCE} in How2Sign. 
Such inconsistencies increase preprocessing overhead and hinder cross-dataset generalization. 
Taken together, these issues point to a lack of interoperability across datasets, highlighting the need for more consistent data formats and unified annotation schemas.

\noindent \textbf{Gloss Quality \& Transferability}
Gloss annotations support recognition and translation by providing structured linguistic supervision, but they remain costly, labor-intensive, and inconsistent in the absence of standardized guidelines. 
Annotator variability, even within a single language (e.g., across German Sign Language corpora), introduces discrepancies that limit effective fine-tuning and cross-corpus transfer. 
At the same time, large-scale datasets such as How2Sign and YouTube-ASL omit gloss annotations entirely, prioritizing scale over structured linguistic grounding. 
Although recent gloss-free approaches have reduced the performance gap, gloss annotations continue to offer advantages in interpretability and training efficiency, particularly in low-resource settings. 
However, inconsistent glossing conventions weaken these benefits by introducing ambiguity in supervision and reducing cross-dataset compatibility. 
As a result, models trained on one dataset often fail to generalize effectively to others. 
Taken together, these observations highlight a fundamental trade-off between annotation quality, consistency, and scalability, pointing to the need for more standardized glossing practices and broader linguistic coverage.

\noindent \textbf{Semantic \& Topical Divergence}
As illustrated in Figure~\ref{fig:three_subfigs}, vocabulary distributions differ substantially across datasets. 
PHOENIX14T primarily reflects weather-related content, whereas How2Sign captures a broader range of instructional scenarios. 
Such domain-specific differences influence SLT performance and limit model generalizability. 
In particular, models trained on narrow-domain corpora often fail to generalize to broader topics without explicit adaptation, indicating a mismatch between training data distributions and real-world usage. 
This issue is further reflected in the separation of semantic representations across datasets, which hinders cross-dataset transfer and joint modeling. 
A more diverse and topic-balanced collection of datasets, especially with consistent gloss annotations, is therefore important for improving robustness in real-world and zero-shot settings. 
In addition, domain-adaptation approaches that exploit semantic relationships across topics offer a promising direction for improving cross-domain generalization. 
Taken together, these observations underscore the need to address semantic fragmentation at both the dataset and modeling levels, while exposing persistent structural deficiencies in availability, coverage, interoperability, supervision consistency, and distribution alignment.

\section{Future Dataset Curation}
\label{sec:future-dataset}
To support scalable and high-quality sign language research, future datasets should prioritize linguistic coverage, ecological realism, multimodal alignment, and interoperable design.
This section distills best practices derived from the challenges and empirical insights discussed earlier.

\noindent \textbf{Video Selection \& Preprocessing}~
To ensure real-world relevance, videos should cover diverse contexts (e.g., greetings, healthcare, education, emergencies, daily life, and news).
Sourcing from open platforms such as YouTube improves topical diversity, but strict filtering is required to remove low-resolution or noisy segments, as fine-grained hand and facial cues are critical.
Datasets should balance sentence length, domain coverage, and linguistic complexity. Long-form videos should be segmented at semantically coherent sign boundaries to eliminate idle frames.
Dataset design must document and balance signer-level attributes, including age, gender, region, dialect, and hand dominance. Hand dominance is particularly important: given the $\sim$10\% prevalence of left-handedness, datasets should actively include left-dominant signers and explicitly report handedness distributions. Where possible, evaluation splits should be stratified by handedness to avoid bias toward right-dominant signing patterns.
Transcriptions (human- or machine-generated) must be verified for temporal alignment and semantic accuracy. To mitigate geographic and dialectal bias, we recommend stratified sampling across regions and dialects, with enforced minimum quotas per group.

\noindent \textbf{Annotation Strategy}~A modular annotation strategy improves usability and extensibility.
At minimum, each video should include a unique identifier and a cleaned sentence-level translation, with additional layers released incrementally.
Gloss annotations provide interpretable intermediates for CSLR and SLT but require expert annotators and are best introduced in later phases.
Temporal sign boundaries, defined by start and end timestamps for each gloss unit, support segmentation and timing-aware generation.
Skeleton-based pose representations are lightweight yet effective across recognition and production, while non-manual cues can be modeled using Facial Action Units (FAUs).
FAUs, derived from the Facial Action Coding System (FACS), encode facial muscle activations with grammatical and affective functions in signed languages~\citep{EkmanFriesenHager2002FACS_RN,zeshan2004interrogative}.
They provide a standardized interface and can be extracted using established toolkits such as OpenFace~\citep{OpenFace}.
This layered strategy enables early data release and later enrichment.

\noindent \textbf{Annotation Tool Selection}~Several tools support sign language annotation.
ELAN~\cite{wittenburg-etal-2006-elan} remains the most widely adopted due to its hierarchical annotation and multimodal support.
Alternative tools such as SignStream and SLAN-tool offer specialized functionality, including linguistic transcription and semi-automated segmentation.
For reference, we summarize their capabilities and limitations in a comparative table in the Appendix.
\section{Conclusion}
We present a survey of 120 sign-language datasets across recognition, translation, and production, identifying key challenges such as uneven coverage, annotation inconsistency, modality imbalance, and fragmented benchmarks. 
Our analysis highlights critical gaps in generalization, evaluation, and cross-dataset comparability, motivating improved dataset design. 
These challenges reflect limitations in coverage, interoperability, and annotation consistency that constrain scalable modeling. 
Overall, this work offers a data-centric perspective connecting datasets, benchmarks, limitations, and design principles for sign-language AI.
\section{Limitations}
\label{sec:limitations}
While our survey offers the most extensive public index of sign-language datasets to date, it is nevertheless subject to six key constraints:
\begin{enumerate}[leftmargin=*]
  \item \textbf{Language imbalance.} Openly available corpora still concentrate on a handful of high-resource sign languages (ASL, DGS, CSL, BSL). Therefore, any conclusions about cross-lingual transferability may fail to generalize to historically under-represented communities—such as many African, Indigenous, and village sign languages—without further evidence.
  \item \textbf{Metadata completeness.}~Statistics such as signer counts were copied verbatim from the original papers or repository \texttt{README}s; we did not re-annotate every clip. Minor inaccuracies may thus persist despite our best cross-checks.
  \item \textbf{Benchmark scope.}~The quantitative leaderboards in Section~\ref{sec:benchmarks} focus on five flagship, general-purpose datasets. Highly specialised domains (e.g., medical or legal signing) remain to be benchmarked in future work.
  \item \textbf{Visualization bias.}~All embedding maps rely on a single UMAP seed and default hyper-parameters.~Alternative random seeds or dimensionality-reduction methods could expose slightly different cluster boundaries.
  \item \textbf{Lack of human evaluation.} We did not yet conduct usability studies with Deaf signers to vet the proposed 24-field datasheet template; structured community feedback therefore remains an essential item on our agenda.
  \item \textbf{Community vetting.}~ We emphasize that the proposed 24-field datasheet is not intended as a finalized compliance standard, but rather as an evolving documentation framework designed to improve transparency and comparability. While it aims to standardize documentation practices, it has not yet been reviewed by Deaf communities or signers, which we consider a key limitation. To address this, we will run an open call for feedback and provide a public GitHub issue template to collect comments; we will summarize and integrate the input in the next iteration and document changes in a public changelog. However, achieving complete and comprehensive validation is challenging: our survey spans 35 sign languages, so participation and collaboration from Deaf communities across these language varieties is essential to make this effort as thorough as possible.
\end{enumerate}

\section*{Broader Impact \& Ethical Considerations}
\label{sec:ethics}
\textbf{Potential benefits.}~By unifying dispersed resources and releasing a standardized datasheet template, we lower entry barriers for newcomers, foster reproducibility, and expose low-resource gaps that merit targeted investment.

\noindent\textbf{Risks and mitigations.} Responsible development of our approach requires careful consideration of potential negative impacts.
\begin{itemize}[leftmargin=*]
\item \emph{Signer privacy.} Many videos display identifiable faces. We therefore urge dataset curators to spell out licence terms and, where appropriate, add options for anonymisation (face-blurring, gated access). See Section~\ref{sec:future-dataset}.
\item \emph{Bias amplification.} Benchmarks dominated by white, Western signers can yield models that under-perform for minority communities. Figure~\ref{fig:world_heatmap} highlights this imbalance; we advocate community-led data collection to correct it.
\item \emph{Malicious use.} Synthetic sign-language output might enable deep-fake content. We recommend visible or invisible watermarks and disclosure when such footage is shared.
\item \emph{Environmental cost.} Our analyses used \textless1 GPU-hour (Appendix B). Still, future large-scale training should report carbon footprints and favour efficient architectures.
  
\end{itemize}

\section*{Acknowledgements}
This work was supported in part by the University of Washington Faculty Startup Fund. 
We thank the anonymous reviewers for their constructive feedback and insightful comments, which helped improve the clarity and scope of this work.

\newpage
\bibliography{anthology}

\newpage

\appendix

\section{Appendix}
This appendix provides supplementary material, including a detailed comparison of annotation tools and comprehensive tables covering the sign language datasets surveyed in this work.

\noindent\textbf{Annotation Tools Details.}
\label{sec:app_tools}
The choice of annotation tools is critical for dataset sustainability, reproducibility, and long-term usability. As discussed in Section~\ref{sec:future-dataset}, several annotation tools are commonly used in sign language research:

\begin{itemize}[leftmargin=1.2em]

    \item \textbf{ELAN}~\cite{wittenburg-etal-2006-elan}  
    ELAN is the most stable and widely adopted annotation platform for sign language corpora. It supports hierarchical tier structures, for example separating gloss annotations from sentence-level translations, and synchronized multimodal streams, including video, audio, and waveform data. Its XML-based storage format facilitates long-term readability and interoperability, making ELAN the preferred choice for large-scale and longitudinal dataset development.
    
    \item \textbf{More Tools}~SignStream~\cite{SignStream} is optimized for fine-grained linguistic transcription of visual–gestural data but offers limited interoperability outside research communities. SLAN-tool~\cite{SLAN} integrates semi-automated neural segmentation to accelerate annotation workflows. However, it depends on ELAN for broader compatibility and may face availability or maintenance constraints.

\end{itemize}

\noindent\textbf{Annotation Tool Comparison.}
Table~\ref{tab:annotation-tools} compares three widely used sign language annotation tools, namely SignStream, ELAN, and SLAN-tool, across dimensions including functionality, usability, interoperability, and modality support within typical sign language research pipelines.

\begin{figure}[h]
    \centering
    \includegraphics[width=0.9\linewidth]{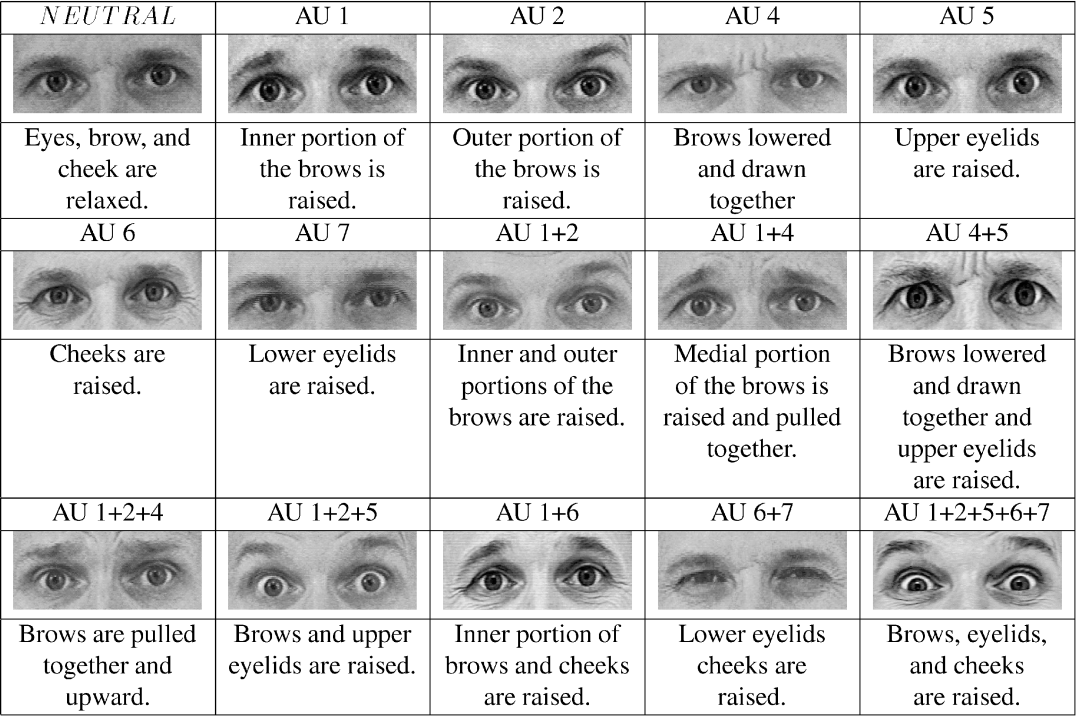}
    \caption{Upper facial Action Units and co-activation patterns. Image adapted from~\cite{tian2001recognizing}.}
    \label{fig:faus}
\end{figure}

\noindent\textbf{Facial Action Units (FAUs).}
The Facial Action Coding System (FACS) provides an anatomically grounded and fine-grained representation of facial expressions by decomposing them into Action Units (AUs), each corresponding to the activation of a specific facial muscle or muscle group. Unlike holistic expression labels, FAUs enable a compositional representation of non-manual signals that are critical to sign language phonology and grammar in practice. Examples include AU1, AU2, and AU4 for eyebrow movement, as well as AU12 for lip-corner activation~\citep{EkmanFriesenHager2002FACS_RN,zeshan2004interrogative,mccullough2009categorical}. Figure~\ref{fig:faus} illustrates several upper facial Action Units, common co-activation patterns, and the facial expressions they convey.

\noindent\textbf{Dataset Tables.}
Tables~\ref{tab:fingerspelling-dataset}--\ref{tab:continuous-dataset-part4} provide a comprehensive overview of the 120 datasets surveyed in this work. The tables are organized by dataset type, including fingerspelling, isolated, and continuous datasets, and summarize key metadata such as sign language, vocabulary size, number of signers, recording modalities, domain coverage, benchmark usage, and evaluation settings across tasks.

However, finer-grained attributes, such as inter-annotator agreement (IAA), annotation guidelines, signer demographics, and recording or collection conditions, are often inconsistently or incompletely reported in the original literature, as discussed in Section~\ref{sec:challenge}. 
To assess transparency, we conduct a dedicated analysis of reporting completeness across both isolated and continuous sign language datasets. Table~\ref{tab:reporting-completeness} summarizes the extent to which these attributes are explicitly documented in the source publications. 
Following a conservative assessment strategy based solely on information explicitly stated in the papers, each attribute is categorized as Covered, Partially Covered, Not Covered, or Unknown. 
Percentages are computed separately over the 54 isolated datasets and the 34 continuous sign language datasets. This analysis focuses on reporting completeness rather than re-evaluating annotation quality or demographic distributions, which are largely unavailable or inconsistently documented in existing sources.

\noindent\textbf{Additional Visualizations.}
Figure~\ref{fig:sentence_Embedding_with_UMAP} presents UMAP projections of sentence-level embeddings for five representative datasets. High-resolution figures and embedding files are archived in the accompanying GitHub repository and are publicly available for further inspection and reuse.

\begin{table}[t]
\centering
\small
\caption{Hand Dominance Reporting Across Surveyed Isolated and Continuous Sign Language Datasets}
\begin{tabular}{lccc}
\hline
Dataset Type & Count & Reported & Not Reported \\
\hline
Isolated & 53 & 5 (9.4\%) & 48 (90.6\%) \\
Continuous & 55 & 5 (9.1\%) & 50 (90.9\%) \\
\hline
Total & 108 & 10 (9.3\%) & 98 (90.7\%) \\
\hline
\end{tabular}
\label{tab:hand_dominance_report}
\end{table}

\noindent\textbf{Hand Dominance Reporting.}
Table~\ref{tab:hand_dominance_report} summarizes the reporting coverage of hand dominance across the surveyed Isolated and Continuous sign language datasets.
Among the 108 datasets included in this survey, only 10 datasets (9.3\%) explicitly report handedness information, while the remaining 98 datasets (90.7\%) do not document this metadata.
The proportion of reporting is nearly identical across dataset types, with 5 out of 53 isolated datasets (9.4\%) and 5 out of 55 continuous datasets (9.1\%) providing explicit hand dominance information.

\begin{table}[t]
\centering
\small
\setlength{\tabcolsep}{4pt}
\caption{Distribution of Sign Language Datasets by Language Resource Level}
\begin{tabular}{lccc}
\hline
Resource Level & Definition & \#Languages & Percentage \\
\hline
High-resource & $\geq$5 datasets & 5 & 14\% \\
Medium-resource & 2--4 datasets & 11 & 31\% \\
Low-resource & 1 dataset & 19 & 54\% \\
\hline
Total & -- & 35 & 100\% \\
\hline
\end{tabular}
\label{tab:low_resource_report}
\end{table}

\noindent\textbf{Low-Resource Language Coverage.}
To examine the distribution of dataset resources across sign languages, we categorize languages based on the number of datasets available for each language, considering both isolated and continuous sign language datasets included in this survey.
As shown in Table~\ref{tab:low_resource_report}, only a small number of sign languages can be considered high-resource, with five or more datasets available.
A limited number of languages fall into the medium-resource category (2--4 datasets), while the majority of sign languages are represented by only a single dataset.

This long-tail distribution indicates that dataset development is heavily concentrated in a small number of well-studied sign languages, such as American Sign Language, Chinese Sign Language, and German Sign Language, while many other languages remain underrepresented.
These results highlight the need for broader dataset collection efforts to support more inclusive and representative sign language technologies.

\begin{table*}[!h]
\centering
\footnotesize
\renewcommand{\arraystretch}{1.08}
\setlength{\tabcolsep}{4pt}
\captionsetup{width=\linewidth}
\caption{\small Comparison of sign language annotation tools across functionality, usability, and integration.}
\label{tab:annotation-tools}
\begin{tabular}{p{2.5cm} >{\raggedright}p{4.1cm} >{\raggedright}p{4.1cm} >{\raggedright\arraybackslash}p{4.1cm}}
\toprule
\textbf{Aspect} & \textbf{SignStream} & \textbf{ELAN} & \textbf{SLAN-tool} \\
\midrule
\textbf{Motivation} 
& Linguistic transcription of visual-gestural languages. 
& Multimodal annotation of natural communication. 
& AI-assisted annotation for sign language NLP. \\

\textbf{Advantages} 
& • Multilevel synchronization \newline • Linguistically detailed annotations 
& • Tier-based structure \newline • Flexible format support \newline • Widely adopted 
& • Neural segmentation \newline • Semi-automatic annotation \newline • ELAN-compatible \\

\textbf{Disadvantages} 
& • Requires expert knowledge \newline • Limited toolchain integration 
& • Steep learning curve \newline • Requires schema familiarity 
& • Dependent on ELAN \newline • Performance tied to pretrained models \\

\textbf{Data Format} 
& Visual-gestural input only; low interoperability 
& Broad audio/video/text support; exportable 
& Optimized for segmentation; integrates with ELAN \\

\textbf{Ease of Use} 
& Researcher-friendly for sign linguists 
& Feature-rich but may require training 
& Customizable GUI for targeted workflows \\

\textbf{Unique Features} 
& Multilevel annotation for both signed and spoken input 
& Timestamped, hierarchical annotation tiers 
& Neural integration for active signing segmentation \\
\bottomrule
\end{tabular}
\end{table*}

% \begin{table*}[htbp]
% \centering
% \small
% \renewcommand{\arraystretch}{1.2}
% \setlength{\tabcolsep}{6pt}
% \caption{Reporting Completeness Across Continuous Sign Language Datasets}
% \label{tab:reporting-completeness}
% \begin{tabular}{lcccc}
% \toprule
% \textbf{Attribute} 
% & \textbf{Covered (\%)} 
% & \textbf{Partially Covered (\%)} 
% & \textbf{Not Covered (\%)} 
% & \textbf{Unknown (\%)} \\
% \midrule
% Inter-Annotator Agreement (IAA)     
% & 0.0  & 8.8  & 82.4 & 8.8  \\
% Annotation Guidelines               
% & 5.9  & 61.8 & 23.5 & 8.8  \\
% Signer Demographics                 
% & 26.5 & 47.1 & 17.6 & 8.8  \\
% Recording / Collection Conditions   
% & 67.6 & 20.6 & 2.9  & 8.8  \\
% \bottomrule
% \end{tabular}
% \end{table*}

\begin{table*}[htbp]
\centering
\small
\renewcommand{\arraystretch}{1.15}
\setlength{\tabcolsep}{5pt}
\caption{Reporting completeness statistics for isolated and continuous sign language datasets.}
\label{tab:reporting-completeness}

\begin{tabular}{lcccccccc}
\toprule
\multirow{2}{*}{\textbf{Attribute}} 
& \multicolumn{4}{c}{\textbf{Isolated (\%)}} 
& \multicolumn{4}{c}{\textbf{Continuous (\%)}} \\
\cmidrule(lr){2-5} \cmidrule(lr){6-9}
& \textbf{Cov.} & \textbf{Part.} & \textbf{Not} & \textbf{Unk.}
& \textbf{Cov.} & \textbf{Part.} & \textbf{Not} & \textbf{Unk.} \\
\midrule
Inter-Annotator Agreement (IAA)
& 5.6  & 3.7  & 79.6 & 11.1
& 0.0  & 8.8  & 82.4 & 8.8  \\
Annotation Guidelines
& 7.4  & 37.0 & 44.4 & 11.1
& 5.9  & 61.8 & 23.5 & 8.8  \\
Signer Demographics
& 18.5 & 57.4 & 13.0 & 11.1
& 26.5 & 47.1 & 17.6 & 8.8  \\
Recording / Collection Conditions
& 66.7 & 22.2 & 0.0  & 11.1
& 67.6 & 20.6 & 2.9  & 8.8  \\
\bottomrule
\end{tabular}

\vspace{2pt}
\parbox{0.98\textwidth}{\footnotesize
\textbf{Cov.} indicates that the attribute is explicitly and clearly documented in the original paper;
\textbf{Part.} indicates that the attribute is mentioned but lacks sufficient detail or completeness;
\textbf{Not} indicates that the attribute is not reported at all;
\textbf{Unk.} indicates that the attribute cannot be reliably determined due to missing or inaccessible information.
}

\end{table*}

\begin{table*}[htbp]
  \centering
  \scriptsize
  \renewcommand{\arraystretch}{1.1}
  \setlength{\tabcolsep}{3pt}
  \caption{FingerSpelling Sign Language Datasets}
  \label{tab:fingerspelling-dataset}
  \resizebox{\textwidth}{!}{%
    \begin{tabular}{%
      p{2.5cm}
      p{0.8cm}
      p{1cm}
      p{1.2cm}
      p{0.8cm}
      p{1.2cm}
      p{0.8cm}
      p{1.5cm}
      p{1.5cm}
      p{1.2cm}
      p{2.2cm} % <<< Hand Dominance 新增
      p{3cm}
      p{0.8cm}
      p{0.8cm}
      p{1.5cm}
    }
    \toprule
    \textbf{Dataset} & \textbf{Year} & \textbf{Language} & \textbf{Vocab.\ Size}
      & \textbf{\#Samples} & \textbf{\#Signers} & \textbf{Domain}
      & \textbf{Collection Source} & \textbf{Resolution}
      & \textbf{Modality} & \textbf{Hand Dominance}
      & \textbf{Publication}
      & \textbf{Available} & \textbf{Task}
      & \textbf{Baseline Model Accuracy} \\
    \midrule

    ChicagoFSWild~\cite{shi2018american}
      & 2018 & American & 31
      & 7,304 sequences & 168 & Letters + Char
      & Online & 640×360 & RGB
      & Right-handed: 6782, Left-handed: 522, Two-hand: 121
      & American Sign Language fingerspelling recognition in the wild
      & \checkmark & SLR & -- \\

    ChicagoFSWild+~\cite{shi2019fingerspelling}
      & 2019 & American & --
      & 55,232 sequences & 260 & Letters + Char
      & Online & -- & RGB
      & Right-handed: 86.9\%, Left-handed: 10.6\%, Other: 2.5\%
      & Fingerspelling recognition in the wild with iterative visual attention
      & \checkmark & SLR & -- \\

    ASL Digits~\cite{mavi2020new}
      & 2020 & American & 10
      & 21,800 images & 218 & Letters
      & Camera & 3024×3024 & RGB
      & --
      & A New Dataset and Proposed Convolutional Neural Network Architecture for Classification of American Sign Language Digits
      & \checkmark & SLR & -- \\

    27 Class ASL~\cite{mavi2022new27classsign}
      & 2022 & American & 27
      & 130 images & 173 & Letters
      & Camera & 3024×3024 & RGB
      & Right Hand Only
      & A New 27 Class Sign Language Dataset Collected from 173 Individuals
      & \checkmark & SLR & -- \\

    FSboard~\cite{georg2025fsboard}
      & 2023 & American & $\sim$3.2M characters
      & 151,000 samples & 147 & Letters
      & Mobile camera & 1944 × 2592
      & RGB Video → Landmark (pose/hand)
      & --
      & FSboard: Over 3 million characters of ASL fingerspelling collected via smartphones
      & \checkmark & SLR
      & 11.1\% CER (52.9\% Top-1 Accuracy, ByT5-small baseline)~\cite{georg2025fsboard}\\

    ArASL~\cite{latif2019arasl}
      & 2019 & Arabic & 32
      & 54,049 images & 40 & Letters
      & Mobile camera & 64×64 & RGB
      & --
      & ArASL: Arabic Alphabets Sign Language Dataset
      & \checkmark & SLR & -- \\

    RGB AASL~\cite{albarham2023rgbarabicalphabetssign}
      & 2023 & Arabic & 31
      & 7,857 images & 200 & Letters
      & Camera & -- & RGB
      & --
      & RGB Arabic Alphabets Sign Language Dataset
      & \checkmark & SLR & -- \\

    AzSLD Fingerspelling~\cite{alishzade2025azsld}
      & 2023 & Azerbaijani & 32
      & 10,864 images, 3,587 videos & 43 & Letters + Gesture
      & Telegram & -- & RGB
      & --
      & AzSLD: Azerbaijani Sign Language Dataset for Fingerspelling, Word, and Sentence Translation with Baseline Software
      & \checkmark & SLR & -- \\

    IsharaKhobor~\cite{rubaiyeat2025banglasignlanguagetranslation}
      & 2012
      & Bangla
      & 37
      & 518 images
      & 3
      & General
      & Lab
      & --
      & RGB Image + Fingertip Position
      & --
      & Bangladeshi Sign Language Recognition using Fingertip Position
      & \checkmark
      & SLR
      & 98.99\% Accuracy \\

    ISL-HS~\cite{oliveira2017dataset}
      & 2017 & Irish & 23
      & 468 videos, 58,114 images & 6 & Letters
      & Mobile camera & 640×480 & RGB
      & --
      & A Dataset for Irish Sign Language Recognition
      & \checkmark & SLR & 95\% Accuracy~\cite{oliveira2017dataset} \\

    RWTH-FingerSpelling~\cite{dreuw2006modeling}
      & 2006 & Germany & 35
      & 1,400 image sequences & 20 & Letters + Umlauts + Number
      & Lab & 320×240, 352×288 & RGB
      & --
      & Modeling Image Variability in Appearance-Based Gesture Recognition
      & \checkmark & SLR & 35.7\% Error Rate~\cite{dreuw2006modeling} \\

    \bottomrule
    \end{tabular}%
  }
\end{table*}

%%%%% Isolated Dataset Table
%---------------- Part I ----------------
\begin{sidewaystable*}[htbp]
  \centering
  \scriptsize
  \renewcommand{\arraystretch}{1.1}
  \setlength{\tabcolsep}{3pt}
  \caption{Isolated Sign Language Dataset (Part I)}
  \label{tab:isolated-dataset-part1}
  \resizebox{0.9\textwidth}{!}{%
    \begin{tabular}{
      p{2.5cm} p{0.8cm} p{1cm}   p{1.2cm} p{0.8cm} p{1.2cm} p{0.8cm}
      p{1.5cm} p{1.5cm} p{1.2cm} p{1.2cm}
      p{1.6cm}  % <<< Hand Dominance 新增
      p{3cm}   p{0.8cm} p{0.8cm} p{1.5cm}
    }
    \toprule
    \textbf{Dataset} & \textbf{Year} & \textbf{Language}
      & \textbf{Vocab.\ Size} & \textbf{Duration} & \textbf{\#Samples}
      & \textbf{\#Signers} & \textbf{Domain} & \textbf{Collection Source}
      & \textbf{Resolution} & \textbf{Modality} & \textbf{Hand Dominance} & \textbf{Publication}
      & \textbf{Available} & \textbf{Task}
      & \textbf{Baseline Model Accuracy} \\
    \midrule
    Alabib-65~\cite{10.1145/3596909}      & 2023 & Algerian
      & 65        & --               & 6,328 videos
      & 29        & General          & iPad Air
      & 720$\times$1,280, 1,080$\times$1,920 & RGB
      & Right-handed: 66.3\%, Left-handed: 5.5\%
      & Alabib-65: A Realistic Dataset for Algerian Sign Language Recognition
      & $\times$           & SLR   & 70.83\%~\cite{10.1145/3596909} \\

    Purdue RVL-SLLL~\cite{inproceedings}    & 2002 & American
      & 101+      & --               & 2,576 video clips
      & 14        & Motion primitives + Handshapes + General & Lab
      & 640×480               & RGB
      & --
      & Purdue RVL-SLLL ASL Database for Automatic Recognition of American Sign Language
      & Contact Author      & SLR   & -- \\

    Boston ASLLVD~\cite{athitsos2008asllvd}       & 2008 & American
      & 3,314     & --               & 9,800 tokens
      & 6         & General          & Lab
      & --                    & RGB
      & --
      & The American Sign Language Lexicon Video Dataset
      & Partially           & SLR   & -- \\

    MSR Gesture3D~\cite{chen2017action}       & 2017 & American
      & 12        & --               & 336 sequences
      & 10        & Gesture          & Lab
      & --                    & RGB-D
      & --
      & Action recognition from depth sequences using weighted fusion of 2D and 3D auto-correlation of gradients features
      & \checkmark         & SLR   & -- \\

    MS-ASL~\cite{joze2018ms}              & 2018 & American
      & 1,000     & $\sim$25 hours  & 25,513 videos
      & 222       & General          & Lab
      & --                    & RGB
      & --
      & MS-ASL: A Large-Scale Data Set and Benchmark for Understanding American Sign Language
      & \checkmark         & SLR, SLP & -- \\

    WLASL~\cite{li2020WLASL}               & 2019 & American
      & 2,000     & $\sim$14 hours  & 21,083 videos
      & 119       & General          & Lab
      & --                    & RGB
      & --
      & Word-level Deep Sign Language Recognition from Video: A New Large-scale Dataset and Methods Comparison
      & \checkmark         & SLR, SLP & Top-10 66.31\% (I3D)~\cite{li2020WLASL} \\

    ASL-100-RGBD~\cite{hassan2020isolated}        & 2020 & American
      & 100       & --               & $\sim$4,150 tokens
      & 22        & General          & Lab
      & 1920$\times$1080, 512$\times$424   & RGB, Skeleton, Depth and HDface
      & --
      & An Isolated-Signing RGBD Dataset of 100 ASL Signs Produced by Fluent ASL Signers
      & \checkmark         & SLR   & - \\

    ASL CrowdSourcing~\cite{bragg2022exploring}   & 2022 & American
      & 60        & --               & 1,906 videos
      & 29        & General          & Crowd
      & --                    & RGB
      & --
      & Exploring Collection of Sign Language Videos through Crowdsourcing
      & $\times$           & SLR   & -- \\

    ASL-Skeleton3D~\cite{de2022asl}      & 2022 & American
      & --        & --               & 9,747 samples
      & 6         & General          & Lab
      & --                    & RGB
      & --
      & ASL-Skeleton3D and ASL-Phono: Two Novel Datasets for the American Sign Language
      & \checkmark         & SLR   & -- \\

    ASL-Phono~\cite{de2022asl}           & 2022 & American
      & 2,294     & --               & 9,747 samples
      & 6         & Linguistics-based & Lab
      & --                    & RGB
      & --
      & ASL-Skeleton3D and ASL-Phono: Two Novel Datasets for the American Sign Language
      & \checkmark         & SLR   & -- \\

    ASLLRP Sign Bank~\cite{neidle2022asl}    & 2022 & American
      & 6,000     & --               & 41,830 lexical signs
      & --        & Lexical           & Lab
      & --                    & RGB
      & --
      & ASL Video Corpora \& Sign Bank: Resources Available through the American Sign Language Linguistic Research Project (ASLLRP)
      & \checkmark         & SLR   & -- \\

    ASL Citizen~\cite{desai2024asl}         & 2023 & American
      & 2,731     & --               & 83,399 videos
      & 52        & General          & Crowd
      & --                    & RGB
      & --
      & ASL Citizen: A Community-Sourced Dataset for Advancing Isolated Sign Language Recognition
      & \checkmark         & SLR   & Top-10 90.86\%~\cite{desai2024asl}  \\

    PopSign ASL v1.0~\cite{starner2023popsign}    & 2024 & American
      & 250       & --               & 214,326 videos
      & 47        & General          & Smartphone
      & --                    & RGB
      & Right-handed: 34, Left-handed: 13
      & PopSign ASL v1.0: An Isolated ASL Dataset Collected via Smartphones
      & Contact Author      & SLR   & 83.80\%~\cite{starner2023popsign} \\

    ArSL corpus~\cite{almohimeed2010arabic}         & 2010 & Arabic
      & 710       & --               & 203 sentences
      & --        & General          & Lab
      & 640×480               & RGB
      & --
      & An Arabic Sign Language corpus for instructional language in school
      & \checkmark         & SLR   & -- \\

    SignsWorld Atlas~\cite{shohieb2015signsworld}    & 2015 & Arabic
      & $\sim$500 & --               & --
      & 10        & General          & Lab
      & --                    & RGB
      & --
      & SignsWorld Atlas; a benchmark Arabic Sign Language database
      & $\times$           & SLR   & -- \\

    LSA-64~\cite{ronchetti2023lsa64}             & 2023 & Argentina
      & 64        & --               & 3,200 video sequences
      & 10        & Dictionary       & Lab
      & --                    & RGB
      & --
      & LSA64: An Argentinian Sign Language Dataset
      & \checkmark         & SLR   & -- \\

    ArSLRS~\cite{ibrahim2018automatic}              & 2018 & Arabic
      & 30        & --               & 450 videos
      & --        & General          & Lab
      & --                    & RGB
      & --
      & An Automatic Arabic Sign Language Recognition System (ArSLRS)
      & $\times$           & SLR   & 97\%~\cite{ibrahim2018automatic} \\

    ArSL for Deaf Drivers~\cite{abbas2021towards}
                        & 2021 & Arabic
      & 215       & --               & 215 videos
      & 3         & Driver           & Lab
      & --                    & RGB
      & --
      & Towards an Arabic Sign Language (ArSL) corpus for deaf drivers
      & \checkmark         & SLR   & 10.23\% WER~\cite{abbas2021towards} \\
    \bottomrule
    \end{tabular}%
  }
\end{sidewaystable*}

%---------------- Part II ----------------
\begin{sidewaystable*}[htbp]
  \centering
  \scriptsize
  \renewcommand{\arraystretch}{1.1}
  \setlength{\tabcolsep}{3pt}
  \caption{Isolated Sign Language Dataset (Part II)}
  \label{tab:isolated-dataset-part2}
  \resizebox{\textwidth}{!}{%
    \begin{tabular}{
      p{2.5cm} p{0.8cm} p{1cm}   p{1.2cm} p{0.8cm} p{1.2cm} p{0.8cm}
      p{1.5cm} p{1.5cm} p{1.2cm} p{1.2cm}
      p{1.6cm}  % <<< Hand Dominance 新增
      p{3cm}   p{0.8cm} p{0.8cm} p{1.5cm}
    }
    \toprule
    \textbf{Dataset} & \textbf{Year} & \textbf{Language}
      & \textbf{Vocab.\ Size} & \textbf{Duration} & \textbf{\#Samples}
      & \textbf{\#Signers} & \textbf{Domain} & \textbf{Collection Source}
      & \textbf{Resolution} & \textbf{Modality} & \textbf{Hand Dominance} & \textbf{Publication}
      & \textbf{Available} & \textbf{Task}
      & \textbf{Baseline Model Accuracy} \\
    \midrule
    KArSL~\cite{sidig2021karsl}          & 2021 & Arabic
      & 502       & --               & 75,300 samples
      & 3         & General          & Lab
      & 1920×1080, 512×424 & RGB-D, Skeleton
      & --
      & KArSL: Arabic Sign Language Database
      & \checkmark         & SLR   & -- \\

    MM-WLAuslan~\cite{shen2024mm}     & 2024 & Australian
      & 3,215     & $\sim$2,500 hours & 282,900 videos
      & 73        & General          & Lab
      & Varies    & RGB-D, Pose data
      & --
      & MM-WLAuslan: Multi-View Multi-Modal Word-Level Australian Sign Language Recognition Dataset
      & \checkmark         & SLR   & - \\

    AzSLD Words~\cite{alishzade2025azsld}    & 2023 & Azerbaijani
      & 100       & --               & --
      & --        & --               & --
      & --        & RGB
      & --
      & AzSLD: Azerbaijani Sign Language Dataset for Fingerspelling, Word \& Sentence Translation with Baseline Software
      & \checkmark         & SLR   & -- \\

    BdSLW60~\cite{Rubaiyeat_2025}
      & 2021 & Bangla
      & 60 & --
      & 9,307 videos
      & 18
      & General
      & Workshop / Lab
      & --
      & Skeleton
      & Right-handed: 7673, Left-handed: 1634
      & BdSLW60: A Word-Level Bangla Sign Language Dataset
      & \checkmark
      & SLR
      & 75.1\% Top-1 Accuracy \\

    BDSL 49~\cite{hasib2023bdsl}        & 2022 & Bangla
      & 49        & --               & 29,490 images
      & 14        & General          & Smartphone
      & --        & RGB
      & --
      & BDSL 49: A Comprehensive Dataset of Bangla Sign Language
      & \checkmark         & SLR   & -- \\

    BdSLW401~\cite{rubaiyeat2025bdslw401transformerbasedwordlevelbangla}
      & 2024 & Bangla
      & 401 & --
      & 102,176 videos
      & 18
      & General
      & Lab
      & --
      & Skeleton
      & Right-handed: 85\%, Left-handed: 15\%
      & BdSLW401: Transformer-Based Word-Level Bangla Sign Language Recognition Using Relative Quantization Encoding (RQE)
      & \checkmark
      & SLR
      & 49.20 WER (Raw, Combined view) \\

    MINDS-Libras    & 2019 & Brazilian
      & 20        & --               & 1,200 videos
      & 12        & Gesture          & Lab
      & 1920×1080 & RGB
      & --
      & (no publication title)
      & \checkmark         & SLR   & -- \\

    BSLDICT~\cite{momeni2020watch}        & 2020 & British
      & 9,283     & --               & 14,210 videos
      & $>$28     & Dictionary       & Website
      & --        & RGB
      & --
      & Watch, read and lookup: learning to spot signs from multiple supervisors
      & \checkmark         & SLR   & -- \\

    DEVISIGN        & 2014 & Chinese
      & 4,414     & --               & 331,050 vocabulary data
      & 30        & General          & Lab
      & --        & RGB-D, Skeleton
      & --
      & (no publication title)
      & Contact Author      & SLR   & -- \\

    CSLR-HMM-D1~\cite{zhang2016chinese}    & 2016 & Chinese
      & 100       & --               & 500 videos
      & 1         & General          & Lab
      & --        & RGB-D, Skeleton
      & --
      & CHINESE SIGN LANGUAGE RECOGNITION WITH ADAPTIVE HMM
      & $\times$           & SLR   & -- \\

    CSLR-HMM-D2~\cite{zhang2016chinese}   & 2016 & Chinese
      & 500       & --               & 2,500 videos
      & 1         & General          & Lab
      & --        & RGB-D, Skeleton
      & --
      & CHINESE SIGN LANGUAGE RECOGNITION WITH ADAPTIVE HMM
      & $\times$           & SLR   & -- \\

    SLR500~\cite{huang2018attention}         & 2018 & Chinese
      & 500       & --               & 125,000 videos
      & 50        & General          & Lab
      & --        & RGB-D, 3D Joints Information
      & --
      & Attention-Based 3D-CNNs for Large-Vocabulary Sign Language Recognition
      & Agreement Needed    & SLR   & 53.8\%~\cite{huang2018attention}\\

    NMFs-CSL~\cite{hu2021global}       & 2020 & Chinese
      & 1,067     & --               & 32,010 videos
      & 10        & General          & Lab
      & --        & RGB
      & --
      & Global-Local Enhancement Network for NMF-Aware Sign Language Recognition
      & Agreement Needed    & SLR   & Top-5 90.5\%~\cite{hu2021global}  \\

    NCSL~\cite{wang20222}           & 2022 & Chinese
      & 300       & --               & 90,000 videos
      & 30        & General          & Lab
      & --        & RGB
      & --
      & (2+1)D-SLR: An Efficient Network for Video Sign Language Recognition
      & $\times$           & SLR   & Top-1 96.4\%~\cite{wang20222} \\

    DGS Kinect 20~\cite{cooper2012sign}  & 2012 & Germany
      & 20        & --               & 840 samples
      & 6         & General          & Lab
      & --        & RGB
      & --
      & Sign Language Recognition Using Sub-Units
      & Contact Author      & SLR   & Top-1 76\%~\cite{cooper2012sign} \\

    DGS Kinect 40~\cite{cooper2012sign}   & 2012 & Germany
      & 40        & --               & 3,000 samples
      & 15        & General          & Lab
      & --        & RGB
      & --
      & Sign Language Recognition Using Sub-Units
      & Contact Author      & SLR   & -- \\

    DW-DGS~\cite{langer2024introducing}         & 2023 & Germany
      & 2,061     & --               & --
      & --        & Dictionary       & Lab
      & --        & RGB
      & --
      & Introducing the DW-DGS – The Digital Dictionary of DGS
      & \checkmark         & SLR   & -- \\

    LSFB-isol~\cite{fink2021lsfb}      & 2021 & French Belgian
      & 395       & --               & 47,551 videos
      & 85        & General          & Lab
      & --        & RGB
      & --
      & LSFB-CONT and LSFB-ISOL: Two New Datasets for Vision-Based Sign Language Recognition
      & \checkmark         & SLR   & Top-1 51.5\%~\cite{fink2021lsfb} \\

    GSL-isol~\cite{adaloglou2021comprehensive}       & 2019 & Greek
      & 310       & 6.44 hours       & 40,785 videos
      & 7         & General          & Lab
      & 840×840   & RGB-D
      & --
      & A Comprehensive Study on Deep Learning-based Methods for Sign Language Recognition
      & \checkmark         & SLR   & 89.74\%~\cite{adaloglou2021comprehensive} \\
    \bottomrule
    \end{tabular}%
  }
\end{sidewaystable*}

%---------------- Part III ----------------
\begin{sidewaystable*}[htbp]
  \centering
  \scriptsize
  \renewcommand{\arraystretch}{1.1}
  \setlength{\tabcolsep}{3pt}
  \caption{Isolated Sign Language Dataset (Part III)}
  \label{tab:isolated-dataset-part3}
  \resizebox{\textwidth}{!}{%
    \begin{tabular}{
      p{2.5cm} p{0.8cm} p{1cm}   p{1.2cm} p{0.8cm} p{1.2cm} p{0.8cm}
      p{1.5cm} p{1.5cm} p{1.2cm} p{1.2cm}
      p{1.6cm}  % <<< Hand Dominance 新增
      p{3cm}   p{1.2cm} p{0.8cm} p{1.5cm}
    }
    \toprule
    \textbf{Dataset} & \textbf{Year} & \textbf{Language}
      & \textbf{Vocab.\ Size} & \textbf{Duration} & \textbf{\#Samples}
      & \textbf{\#Signers} & \textbf{Domain} & \textbf{Collection Source}
      & \textbf{Resolution} & \textbf{Modality} & \textbf{Hand Dominance} & \textbf{Publication}
      & \textbf{Available} & \textbf{Task}
      & \textbf{Baseline Model Accuracy} \\
    \midrule

    IISL Nandy 2010~\cite{nandy2010recognition} & 2010 & Indian
      & 22        & --               & 600 samples
      & --        & General          & Lab
      & --        & RGB
      & --
      & Recognition of Isolated Indian Sign Language Gesture in Real Time
      & $\times$           & SLR   & -- \\

    INSLR Dataset~\cite{kishore2012video}     & 2012 & Indian
      & 80        & --               & 1,600 videos
      & 10        & General          & Lab
      & 640×480   & RGB
      & --
      & A Video Based Indian Sign Language Recognition System (INSLR) Using Wavelet Transform and Fuzzy Logic
      & $\times$           & SLR   & 96\%~\cite{kishore2012video} \\

    INCLUDE~\cite{sridhar2020include}          & 2020 & Indian
      & 263       & --               & 4,287 videos
      & 7         & General          & Lab
      & 1920×1080 & RGB
      & --
      & INCLUDE: A Large Scale Dataset for Indian Sign Language Recognition
      & \checkmark         & SLR   & -- \\

    CISLR~\cite{joshi2022cislr}            & 2022 & Indian
      & 4,765     & --               & 7,050 videos
      & 71        & General          & Lab
      & --        & RGB
      & --
      & CISLR: Corpus for Indian Sign Language Recognition
      & Agreement Needed    & SLR   & -- \\

    IISL2020~\cite{kothadiya2022deepsign}        & 2022 & Indian
      & 11        & --               & $\sim$12,100 videos
      & 16        & General          & Lab
      & 1920×1080 & RGB
      & --
      & Deepsign: Sign Language Detection and Recognition Using Deep Learning
      & \checkmark         & SLR   & F1-Score 97\%~\cite{kothadiya2022deepsign} \\

    K-RSL~\cite{mukushev2020evaluation}            & 2020 & Kazakh-Russian
      & 20        & --               & 5,200 isolated sign samples
      & 5         & General          & Lab
      & --        & RGB, Skeleton-Keypoints
      & --
      & Evaluation of Manual and Non-manual Components for Sign Language Recognition
      & \checkmark         & SLR   & 78.20\%~\cite{mukushev2020evaluation} \\

    KSL-Dataset~\cite{yang2019korean}      & 2019 & Korean
      & 77        & --               & 1,229 videos
      & 22        & General          & Lab
      & 255×255   & RGB
      & --
      & The Korean Sign Language Dataset for Action Recognition
      & $\times$           & SLR   & -- \\

    KSL Shin 2023~\cite{shin2023korean}   & 2023 & Korean
      & 20        & $\sim$1,600 seconds    & 400 videos
      & 20        & General          & Lab
      & --        & RGB
      & --
      & Korean Sign Language Recognition Using Transformer-Based Deep Neural Network
      & $\times$           & SLR   & 98.30\%~\cite{shin2023korean} \\

    MSL~\cite{mejia2022automatic}              & 2022 & Mexican
      & 30        & --               & 3,000 samples
      & 4         & General          & Lab
      & 4056×3040, 1280×800 & RGB-D
      & --
      & Automatic Recognition of Mexican Sign Language Using a Depth Camera and Recurrent Neural Networks
      & \checkmark         & SLR   & 96.44\%~\cite{mejia2022automatic} \\

    WLPSL             & --   & Pakistani
      & 31        & --               & 248 videos
      & 12        & General          & Lab
      & --        & RGB
      & --
      & WLPSL: Word-Level Pakistani Sign Language Video Dataset
      & \checkmark         & SLR   & -- \\

    PSL-30~\cite{oszust2013polish}           & 2013 & Polish
      & 30        & --               & 300 videos
      & 1         & General          & Lab
      & 640×480   & RGB-D, Skeleton
      & --
      & Polish Sign Language Words Recognition with Kinect
      & $\times$           & SLR   & Top-1 98.33\%~\cite{oszust2013polish} \\

    KSU-SSL~\cite{al2020hand}          & 2020 & Saudi
      & 40        & --               & --
      & --        & General          & Lab
      & Varies    & RGB, Kinect
      & --
      & Hand Gesture Recognition for Sign Language Using 3DCNN
      & $\times$           & SLR   & -- \\

    LSE-Sign~\cite{gutierrez2016lse}        & 2015 & Spanish
      & 5,100     & --               & 5,100 entries
      & 2         & Dictionary       & Lab
      & --        & RGB
      & --
      & LSE-Sign: A lexical database for Spanish Sign Language
      & Agreement Needed    & SLR   & -- \\

    SL-Animals-DVS~\cite{vasudevan2020introduction}   & 2020 & Spanish
      & 19        & --               & 1,102 recordings
      & 58        & Animal           & YouTube
      & 128×128   & RGB
      & --
      & Introduction and Analysis of an Event-Based Sign Language Dataset
      & \checkmark   & SLR   & -- \\

    SSL Lexicon~\cite{mesch2012meaning}      & 2012 & Swedish
      & 21,345    & --               & --
      & --        & General          & Lab
      & --        & RGB
      & --
      & From meaning to signs and back: Lexicography and the Swedish Sign Language Corpus
      & \checkmark         & SLR   & -- \\

    SMILE~\cite{ebling2018smile}            & 2018 & Swiss-German
      & 100       & --               & --
      & 30        & General          & Lab
      & Varies    & RGB-D
      & --
      & SMILE Swiss German Sign Language Dataset
      & \checkmark         & SLR   & -- \\

    BosphorusSign~\cite{camgoz2016bosphorussign}    & 2016 & Turkish
      & 855       & --               & --
      & 10        & Health, Finance, General & Lab
      & 1920×1080 & RGB-D
      & --
      & BosphorusSign: A Turkish Sign Language Recognition Corpus in Health and Finance Domains
      & $\times$           & SLR   & -- \\

    BosphorusSign22k~\cite{ozdemir2020bosphorussign22k}  & 2020 & Turkish
      & 744       & $\sim$19 hours  & 22,542 videos
      & 6         & Health, Finance, General & Lab
      & 1920×1080 & RGB-D
      & --
      & BosphorusSign22k Sign Language Recognition Dataset
      & Contact Author      & SLR   & Top-5 94.76\%~\cite{ozdemir2020bosphorussign22k} \\

    AUTSL~\cite{sincan2020autsl}             & 2020 & Turkish
      & 226       & 21 hours        & 38,336 samples
      & 43        & General          & Lab
      & 512×512   & RGB-D, Skeleton
      & Right-handed: 41, Left-handed: 2
      & AUTSL: A Large Scale Multi-Modal Turkish Sign Language Dataset and Baseline Methods
      & \checkmark         & SLR   & Top-5 83.93\%~\cite{sincan2020autsl}  \\
    \bottomrule
  \end{tabular}%
  }
\end{sidewaystable*}

%---------------- Continuous Part I ----------------
\begin{sidewaystable*}[htbp]
  \centering
  \scriptsize
  \renewcommand{\arraystretch}{1.1}
  \setlength{\tabcolsep}{3pt}
  \caption{Continuous Sign Language Datasets (Part I)}
  \label{tab:continuous-dataset-part1}
  \resizebox{\textwidth}{!}{%
    \begin{tabular}{
      p{2.5cm} p{0.8cm} p{1cm}   p{1.2cm} p{0.8cm} p{1.2cm} p{0.8cm}
      p{1.5cm} p{1.5cm} p{1.2cm} p{1.2cm}
      p{1.6cm}  % <<< Hand Dominance 新增
      p{3cm}   p{0.8cm} p{0.8cm} p{1.5cm}
    }
    \toprule
    \textbf{Dataset} & \textbf{Year} & \textbf{Language}
      & \textbf{Vocab.\ Size} & \textbf{Duration} & \textbf{\#Samples}
      & \textbf{\#Signers} & \textbf{Domain} & \textbf{Collection Source}
      & \textbf{Resolution} & \textbf{Modality} & \textbf{Hand Dominance} & \textbf{Publication}
      & \textbf{Available} & \textbf{Task}
      & \textbf{Baseline Model Accuracy} \\
    \midrule

    RWTH‐Boston‐104~\cite{dreuw2007speech}       & 2007 & American & 104      & 8.7 min     & 201 sent.    & 3   & General   & Lab
      & –           & RGB
      & --
      & Speech Recognition Techniques for a Sign Language Recognition System
      & \checkmark & SLR      & 17\% WER~\cite{dreuw2007speech} \\

    RWTH‐Boston‐400       & 2008 & American & $\sim$400& –           & 843 sent.    & 5   & General   & Lab
      & –           & RGB
      & --
      & –
      & $\times$   & SLR      & –          \\

    CopyCat~\cite{zafrulla2010novel}               & 2010 & American & 22       & –           & 420 phrases  & 5   & General   & Lab
      & –           & RGB
      & Right-Handed only
      & A novel approach to ASL Phrase Verification using Reversed Signing
      & $\times$   & SLR      & –          \\

    NCSLGR~\cite{neidle2012new}                & 2012 & American & 1{,}920  & –           & 1{,}887 utt. & 4   & General   & Lab
      & –           & RGB
      & --
      & A New Web Interface to Facilitate Access to Corpora
      & \checkmark & SLR      & –          \\

    ASLG‐PC12~\cite{othman2012english}             & 2012 & American & –        & –           & 100 M sent.  & –   & General   & Lab
      & –           & RGB
      & --
      & English–ASL Gloss Parallel Corpus 2012: ASLG‐PC12
      & \checkmark & SLR      & –          \\

    How2Sign~\cite{duarte2021how2sign}              & 2020 & American & 16{,}000 & 79 hours        & >35{,}000 sent.& 11 & General   & Lab
      & 1280×720    & RGB, RGB-D, 3D Keypoints
      & --
      & How2Sign: A Large-scale Multimodal Dataset for Continuous American Sign Language
      & \checkmark & SLR, SLT, SLP & – \\

    ASLing~\cite{ananthanarayana2021dynamic}                & 2021 & American & –        & –           & 1{,}284 samples & 7 & General   & Crowd
      & 450×600     & RGB
      & --
      & Dynamic Cross-Feature Fusion for American Sign Language Translation
      & $\times$   & SLT      & –          \\

    OpenASL~\cite{shi2205open}               & 2022 & American & 33{,}000 & 288 hours       & –            & ~220& General   & Web
      & –           & RGB
      & Right-handed: 83\%, Left-handed: 17\%
      & Open-Domain Sign Language Translation Learned from Online Video
      & \checkmark & SLT      & BLEU$_4$ 6.72~\cite{shi2205open} \\

    ASL‐Homework‐RGBD~\cite{hassan2022asl}     & 2022 & American & –        & –           & 935 samples  & 45  & General   & Homework
      & –           & RGB-D
      & --
      & ASL-Homework-RGBD Dataset: 45 signers’ ASL homework videos
      & \checkmark & SLT      & –          \\

    YouTube‐ASL~\cite{uthus2024youtube}           & 2023 & American & 60{,}000 & $\sim$1000 hours & –            & >2{,}500& General & Web
      & –           & RGB
      & --
      & YouTube‐ASL: A Large‐Scale, Open‐Domain ASL–English Parallel Corpus
      & \checkmark & SLT      & BLEU$_4$ 3.95~\cite{uthus2024youtube} \\

    DailyMoth‐70h~\cite{rust2024towards}         & 2024 & American & 19{,}694 & 75.8 hours      & 48{,}386 clips& 1   & News      & TV
      & –           & RGB
      & --
      & Towards Privacy-Aware Sign Language Translation at Scale
      & \checkmark & SLT      & BLEU$_4$ 28.8~\cite{rust2024towards} \\

    Auslan‐Daily Comm.~\cite{shen2024auslan}    & 2024 & Australian& 3{,}064  & –           & 14{,}041 sent.& 49  & General   & TV\,/\,Web
      & 1920×1080   & RGB
      & --
      & Auslan-Daily: Australian SLT for Daily Communication and News
      & \checkmark & SLT      & BLEU$_4$ 9.95~\cite{shen2024auslan} \\

    Auslan‐Daily News~\cite{shen2024auslan}     & 2024 & Australian& 12{,}346 & –           & 11{,}065 sent.& 18  & General   & TV\,/\,Web
      & 1280×720, 1920×1080 & RGB
      & --
      & Auslan-Daily: Australian SLT for Daily Communication and News
      & \checkmark & SLT      & BLEU$_4$ 2.81~\cite{shen2024auslan} \\

    BTVSL~\cite{zeeon2024btvsl}                 & 2024 & Bangla    & 48{,}623 & 60 hours        & 24{,}085 sent.& 22  & News      & Web
      & –           & RGB
      & --
      & BTVSL: A Novel Sentence-Level Annotated Dataset for Bangla SLT
      & $\times$   & SLT      & BLEU$_4$ 25.16~\cite{zhou2023glossfreesignlanguagetranslation, zeeon2024btvsl} \\

    LIBRAS‐UFOP           & 2021 & Brazilian & 56       & –           & 3{,}040 seq.  & 5   & General   & Lab
      & –           & RGB, RGB-D, 3D Keypoints
      & --
      & A multimodal LIBRAS-UFOP dataset of minimal pairs
      & $\times$   & SLR      & –          \\

    BSL-1K~\cite{albanie2020bsl}               & 2020 & British   & 1{,}064  & $\sim$1000 hours & 273{,}000 sam.& 40  & General   & TV
      & –           & RGB
      & --
      & BSL-1K: Scaling up co-articulated SLR using mouthing cues
      & \checkmark & SLR      & Top-5 88.83\%~\cite{albanie2020bsl} \\
      
    \bottomrule
    \end{tabular}%
  }
\end{sidewaystable*}

%---------------- Continuous Part II ----------------
\begin{sidewaystable*}[htbp]
  \centering
  \scriptsize
  \renewcommand{\arraystretch}{1.1}
  \setlength{\tabcolsep}{3pt}
  \caption{Continuous Sign Language Datasets (Part II)}
  \label{tab:continuous-dataset-part2}
  \resizebox{\textwidth}{!}{%
    \begin{tabular}{
      p{2.5cm} p{0.8cm} p{1cm}   p{1.2cm} p{0.8cm} p{1.2cm} p{0.8cm}
      p{1.5cm} p{1.5cm} p{1.2cm} p{1.2cm}
      p{1.6cm}  % <<< Hand Dominance 新增
      p{3cm}   p{1.2cm} p{0.8cm} p{1.5cm}
    }
    \toprule
    \textbf{Dataset} & \textbf{Year} & \textbf{Language}
      & \textbf{Vocab.\ Size} & \textbf{Duration} & \textbf{\#Samples}
      & \textbf{\#Signers} & \textbf{Domain} & \textbf{Collection Source}
      & \textbf{Resolution} & \textbf{Modality} & \textbf{Hand Dominance} & \textbf{Publication}
      & \textbf{Available} & \textbf{Task}
      & \textbf{Baseline Model Accuracy} \\
    \midrule

    BOBSL~\cite{albanie2021BBC_Oxford}                 & 2021 & British   & 2{,}281/78{,}000 & 1{,}467 hours & 1.2 M seq.& 39  & General   & TV
      & –           & RGB
      & --
      & BBC-Oxford British Sign Language Dataset
      & \checkmark & SLR, SLT & –          \\

    Video-based CSL~\cite{huang2018video}       & 2018 & Chinese   & 178      & 100+ hours      & 25{,}000 inst.& 50  & General   & Lab
      & 1920×1080   & RGB-D
      & --
      & Video-based Sign Language Recognition without Temporal Segmentation
      & $\times$   & SLR      & –          \\

    CSLD~\cite{yuan2019large}                & 2019 & Chinese   & 10{,}000 & –           & 49{,}708 vid.& 50  & General   & Lab
      & 1920×1080, 512×424 & RGB-D
      & --
      & Large Scale Sign Language Interpretation
      & Contact Author & SLR & BLEU$_1$ 14.28~\cite{yuan2019large}  \\

    CSL-Daily~\cite{zhou2021CSLDaily}           & 2021 & Chinese   & 2{,}000  & –           & 20{,}645 vid.& 10  & General   & Lab
      & 1920×1080   & RGB
      & --
      & Improving SLT with Monolingual Data by Sign Back-Translation
      & Agreement Needed & SLR, SLT & BLEU$_4$ 21.34~\cite{zhou2021CSLDaily} \\

    CSL-News~\cite{li2025unisignunifiedsignlanguage} &
    2025 &
    Chinese &
    4{,}875 &
    1{,}985 hours &
    751{,}320 pairs &
    -- &
    News &
    TV &
    Vary &
    RGB &
    -- &
    Uni-Sign: Toward Unified Sign Language Understanding at Scale &
    $\times$ &
    SLT &
    -- \\

    CoL-SLTD~\cite{rodriguez2020understanding}             & 2020 & Colombian & –        & –           & 1{,}020 vid. & 13  & General   & Lab
      & 448×448     & RGB
      & --
      & Understanding Motion in Sign Language: A New Structured Translation Dataset
      & –          & SLT      & –          \\

    S-pot~\cite{viitaniemi2014s}               & 2014 & Finnish   & 1{,}211  & –           & 5{,}539 vid. & 5   & General   & Lab
      & 720×576     & RGB
      & Right-handed: 4, Left-handed: 1
      & S-pot: A benchmark in spotting signs within continuous signing
      & Contact Author & SLR & 47.70\%~\cite{viitaniemi2014s} \\

    VRT-NEWS~\cite{camgoz2021content4all}             & 2021 & Flemish   & 6{,}875  & $\sim$9 hours   & 7{,}174 seq. & 9   & News      & TV
      & 1280×720    & RGB
      & --
      & Content4All Open Research SLT Datasets
      & \checkmark & SLT      & BLEU$_4$ 0.36~\cite{camgoz2021content4all} \\

    Corpus VGT            & –    & Flemish   & –        & 140 hours       & –            & 120 & General   & Lab
      & –           & RGB
      & --
      & –
      & \checkmark & –        & –          \\

    Mediapi-RGB~\cite{ouakrim2024mediapi}          & 2024 & French    & 27{,}343 & 86 hours        & 1{,}230 vid.& >10 & General   & Online Media
      & Vary        & RGB
      & --
      & Mediapi-RGB: An extensive LSF video–text corpus
      & \checkmark & SLT      & BLEU$_4$ 4.14~\cite{ouakrim2024mediapi}  \\

     LSFB-CONT~\cite{fink2021lsfb}         & 2021 & French Belgian & 6{,}883 & --
      & 85{,}132 videos & 100  & General  & Lab
      & --
      & Right-handed: 74.0\%, Left-handed: 14.0\%, Ambidextrous: 7.0\%, Unknown: 5.0\%
      & LSFB-CONT and LSFB-ISOL: Two New Datasets for Vision-Based Sign Language Recognition
      & \checkmark  & --       & -- \\

    SIGNUM~\cite{von2010signum}           & 2008 & Germany      & 450      & 55.3 hours
      & 33{,}210 seq.  & 25   & General  & Lab
      & 776$\times$578
      & Right-handed: 23, Left-handed: 2
      & SIGNUM Database: Video Corpus for Signer-Independent Continuous SL Recognition
      & \checkmark  & SLR      & -- \\

    RWTH-PHOENIX 2012~\cite{forster2012rwth}  & 2012 & Germany      & 911      & 3.25 hours
      & 1{,}980 sent.  & 7    & Weather  & TV
      & 210$\times$260
      & --
      & RWTH-PHOENIX-Weather: A large‐vocabulary SL recognition \& translation corpus
      & \checkmark  & SLR/SLT  & -- \\
    \bottomrule
    \end{tabular}%
  }
\end{sidewaystable*}

%---------------- Continuous Part III ----------------
\begin{sidewaystable*}[htbp]
  \centering
  \scriptsize
  \renewcommand{\arraystretch}{1.1}
  \setlength{\tabcolsep}{3pt}
  \caption{Continuous Sign Language Datasets (Part III)}
  \label{tab:continuous-dataset-part3}
  \resizebox{\textwidth}{!}{%
    \begin{tabular}{
      p{2.5cm} p{0.8cm} p{1cm} p{1.2cm} p{0.8cm}
      p{1.2cm} p{0.8cm} p{1.2cm} p{1.2cm}
      p{1.2cm}  % Resolution
      p{2.2cm}  % Hand Dominance
      p{3cm}
      p{0.8cm} p{1cm} p{1cm}
    }
    \toprule
    \textbf{Dataset} & \textbf{Year} & \textbf{Language}
      & \textbf{Vocab.\ Size} & \textbf{Duration}
      & \textbf{\#Samples} & \textbf{\#Signers}
      & \textbf{Domain} & \textbf{Collection Source}
      & \textbf{Resolution}
      & \textbf{Hand Dominance}
      & \textbf{Publication}
      & \textbf{Available} & \textbf{Task}
      & \textbf{Baseline Acc.} \\
    \midrule

    RWTH-PHOENIX 2014~\cite{forster2014extensions} & 2014 & Germany      & 1{,}558  & 10.73 hours
      & 6{,}861 sent.  & 9    & Weather  & TV
      & 210$\times$260
      & --
      & Extensions of the Sign Language Recognition \& Translation Corpus RWTH-PHOENIX-Weather
      & \checkmark  & SLR/SLT  & -- \\

    Public DGS Corpus~\cite{jahn2018publishing}  & 2018 & Germany      & --
      & $>$50 hours & -- & 327  & General  & Lab
      & 640$\times$360
      & --
      & Publishing DGS corpus data: Different Formats for Different Needs
      & \checkmark  & --       & -- \\

    RWTH-PHOENIX14T~\cite{camgoz2020sign}    & 2020 & Germany      & 2{,}887  & $\sim$10.5 hours
      & 8{,}257 sent.  & 9    & Weather  & TV
      & 210$\times$260
      & --
      & Sign Language Transformers: Joint End-to-end SL Recognition \& Translation
      & \checkmark  & SLR/SLT  & WER 26.5~\cite{koller2019weakly},\,BLEU$_4$ 9.58~\cite{camgoz2018neural} \\

    SWISSTXT-WEATHER~\cite{camgoz2021content4all}   & 2021 & Germany      & 1{,}248  & $\sim$1 hours
      & 811 seq.  & 1    & Weather  & TV
      & 1280$\times$720
      & --
      & Content4All Open Research SLT Datasets
      & \checkmark  & --       & -- \\

    SWISSTXT-NEWS~\cite{camgoz2021content4all}     & 2021 & Germany      & 10{,}561 & $\sim$9.5 hours
      & 6{,}031 seq.  & 9    & News  & TV
      & 1280$\times$720
      & --
      & Content4All Open Research SLT Datasets
      & \checkmark  & SLT      & BLEU$_4$\,0.41~\cite{camgoz2021content4all} \\

    PHOENIX-News~\cite{yin2024t2sgptdynamicvectorquantization}  & 2024 & Germany
      & 190{,}000 & 486 hours & -- & 11 & News & TV
      & --
      & --
      & T2S-GPT: Dynamic Vector Quantization for Autoregressive SL Production from Text
      & Contact Author & SLP & -- \\

    GRSL & 2020 & Greek & $\geq$1{,}500 & -- 
      & $\geq$4{,}000 sent. & $\geq$15 & General & Lab
      & 1920$\times$1080, 1232$\times$1028, 512$\times$524
      & --
      & Towards a visual Sign Language dataset for home care services
      & -- & -- & -- \\

    GSL SD & 2021 & Greek & 310 & 9.59 hours
      & 10{,}295 videos & 7 & General & Lab
      & 848$\times$480
      & --
      & A Comprehensive Study on Deep Learning-based Methods for Sign Language Recognition
      & \checkmark & -- & -- \\

    GSL SI & 2021 & Greek & 310 & 9.59 hours
      & 10{,}295 videos & 7 & General & Lab
      & 848$\times$480
      & --
      & A Comprehensive Study on Deep Learning-based Methods for Sign Language Recognition
      & \checkmark & -- & -- \\

    Elementary23 & 2023 & Greek & 23{,}204 & 71 hours
      & 29{,}653 videos & 9 & General & Lab
      & 1280$\times$720
      & --
      & A New Dataset for End-to-End Sign Language Translation: The Greek Elementary School Dataset
      & $\times$ & SLT & BLEU$_4$ 6.67 \\

    TVB-HKSL-News & 2024 & Hong Kong & SLR 6{,}515, SLT 2{,}850 & 16.07 hours
      & 7k videos & 2 & News & TV
      & 248$\times$360
      & --
      & A Hong Kong Sign Language Corpus Collected from Sign-interpreted TV News
      & Contact Author & SLR, SLT & WER 34.08\%, BLEU$_4$ 23.58 \\
      
    \bottomrule
    \end{tabular}%
  }
\end{sidewaystable*}

%---------------- Continuous Part IV ----------------
\begin{sidewaystable*}[htbp]
  \centering
  \scriptsize
  \renewcommand{\arraystretch}{1.1}
  \setlength{\tabcolsep}{3pt}
  \caption{Continuous Sign Language Datasets (Part IV)}
  \label{tab:continuous-dataset-part4}
  \resizebox{\textwidth}{!}{%
    \begin{tabular}{
      p{2.5cm} p{0.8cm} p{1cm} p{1.2cm} p{0.8cm}
      p{1.2cm} p{0.8cm} p{1.2cm} p{1.2cm}
      p{1.2cm}  % Resolution
      p{2.2cm}  % Hand Dominance
      p{3cm}
      p{0.8cm} p{1cm} p{1cm}
    }
    \toprule
    \textbf{Dataset} & \textbf{Year} & \textbf{Language}
      & \textbf{Vocab.\ Size} & \textbf{Duration}
      & \textbf{\#Samples} & \textbf{\#Signers}
      & \textbf{Domain} & \textbf{Collection Source}
      & \textbf{Resolution}
      & \textbf{Hand Dominance}
      & \textbf{Publication}
      & \textbf{Available} & \textbf{Task}
      & \textbf{Baseline Acc.} \\
    \midrule

    ISL-CSLTR & 2021 & Indian & -- & --
      & 700 videos & 7 & General & Lab
      & --
      & --
      & --
      & \checkmark & SLR, SLT & -- \\

    ISLTranslate & 2023 & Indian & 11k & --
      & 31k & -- & General & DEF, ISLRTC
      & --
      & --
      & ISLTranslate: Dataset for Translating Indian Sign Language
      & \checkmark & SLT & BLEU$_4$ 6.09 \\

    iSign & 2024 & Indian & 40k & 252 hours
      & 118k sent. & -- & General & Web, News
      & --
      & --
      & iSign: A Benchmark for Indian Sign Language Processing
      & \checkmark & SLR, SLT & Top-5 20.04\%, BLEU$_4$ 1.47 \\

    Deep JSLC & 2018 & Japanese & 197 & --
      & 931 sent. & 1 & General & Lab
      & --
      & --
      & Deep JSLC: A Multimodal Corpus Collection for Data-driven Generation of Japanese Sign Language Expressions
      & $\times$ & SLP & -- \\

    KETI & 2018 & Korean & 524 & 20.05 hours
      & 14{,}672 videos & 14 & Emergency & Lab
      & --
      & --
      & Neural Sign Language Translation based on Human Keypoint Estimation
      & $\times$ & SLT & 55.28\% \\

    Corpus NGT & 2008 & Netherlands & 3{,}300 & 12 hours
      & 160 videos & 100 & General & Lab
      & --
      & --
      & The Corpus NGT: an online corpus for professionals and laymen
      & \checkmark & SLR & -- \\

    RKS-PERSIANSIGN & 2020 & Persian & 100 & --
      & 10{,}000 videos & 10 & General & Lab
      & --
      & --
      & Hand sign language recognition using multi-view hand skeleton
      & $\times$ & SLR & 99.80\% \\

    PeruSIL & 2022 & Peruvian & $>$500 & --
      & $>$150 sent. & -- & General & Web
      & --
      & --
      & PeruSIL: A Framework to Build a Continuous Peruvian Sign Language Interpretation Dataset
      & \checkmark & SLR & -- \\

    TheRuSLan & 2020 & Russian & 164 & $>$8 hours
      & $>$10{,}660 samples & 13 & Supermarket & Lab
      & --
      & --
      & TheRuSLan Dataset
      & \checkmark & SLR & -- \\

    Slovo & 2023 & Russian & 1{,}000 & 19.81 hours
      & 20{,}000 videos & 194 & General & Crowdsourcing
      & --
      & --
      & Slovo: Russian Sign Language Dataset
      & \checkmark & SLR & -- \\

    LSA-T & 2022 & Spanish & 14{,}239 & 21.78 hours
      & 14{,}880 sent. & 103 & General & Web
      & 1920$\times$1080
      & --
      & LSA-T: The first continuous Argentinian Sign Language dataset for SLT
      & \checkmark & SLT & -- \\

    SSLC & 2012 & Swedish & 3{,}600 & --
      & 42 videos & 42 & General & Lab
      & --
      & --
      & Sign Language Resources in Sweden : Dictionary and Corpus
      & Partially & SLR & -- \\

    STS-korpus & 2020 & Swedish & -- & --
      & -- & 42 & Teaching & Lab
      & 768$\times$288
      & --
      & STS-korpus: A Sign Language Web Corpus Tool for Teaching and Public Use
      & Free to visit & -- & -- \\

    ATIS & 2008 & Multi & -- & --
      & 595 sent. & -- & General & Lab
      & --
      & --
      & The ATIS Sign Language Corpus
      & $\times$ & -- & -- \\

    Dicta-Sign & 2012 & Multi & $\sim$1{,}000 & --
      & -- & 14--16/lang & General & --
      & --
      & --
      & Dicta-Sign – Building a Multilingual Sign Language Corpus
      & $\times$ & -- & -- \\

    AFRISIGN & 2023 & Multi & 20k & 152 hours
      & -- & -- & General & Web
      & --
      & --
      & AFRISIGN: Machine Translation for African Sign Languages
      & $\times$ & -- & -- \\
    \bottomrule
    \end{tabular}%
  }
\end{sidewaystable*}

\end{document}